\pdfoutput=1
\documentclass[11pt]{article}

\usepackage[final]{acl}
\usepackage{times}
\usepackage{latexsym}
\usepackage[T1]{fontenc}
\usepackage[utf8]{inputenc}
\usepackage{microtype}
\usepackage{inconsolata}
\usepackage{graphicx}
\usepackage[utf8]{inputenc} 
\usepackage[T1]{fontenc}   
\usepackage{hyperref}     
\usepackage{url}            
\usepackage{booktabs}     
\usepackage{amsfonts}       
\usepackage{nicefrac}      
\usepackage{microtype}       
\usepackage{xcolor}        
\usepackage{wrapfig}
\usepackage{longtable}
\usepackage{booktabs}
\usepackage{wrapfig}
\usepackage{lipsum} 
\usepackage{booktabs}
\usepackage{caption}
\usepackage{url}        
\usepackage{booktabs}   
\usepackage{amsfonts}   
\usepackage{nicefrac}   
\usepackage{microtype}  
\usepackage{xcolor}     
\usepackage{graphicx}
\usepackage{tcolorbox}
\usepackage{amsmath}
\usepackage[utf8]{inputenc}
\usepackage{siunitx}
\usepackage{float}
\usepackage{booktabs}
\usepackage{array} 
\usepackage{placeins}

\usepackage{cuted}
\usepackage{caption}
\usepackage{dblfloatfix}

\usepackage{subcaption}
\definecolor{average}{HTML}{006101}
\definecolor{ideal}{HTML}{286ba8}
\definecolor{sample}{HTML}{ff3f37}

\usepackage{tabularx}
\usepackage{wrapfig}
\usepackage{enumitem}
\newcommand{\etal}{\textit{et al}. }

\title{A Theory of Response Sampling in LLMs: \\Part \textcolor{average}{Descriptive} and Part \textcolor{ideal}{Prescriptive}}

\author{{$^{*}$Sarath Sivaprasad$^{1}$, $^{*}$Pramod Kaushik$^{2}$, Sahar Abdelnabi$^{3}$, Mario Fritz$^{1}$} \\
  {\small
  $^{1}$CISPA Helmholtz Center for Information Security \quad
  $^{2}$TCS Research, Pune \quad
  $^{3}$Microsoft} \\
  {\small
  \texttt{\{sarath.sivaprasad, fritz\}@cispa.de} \quad
  \texttt{pramod.kaushik@tcs.com} \quad
  \texttt{saabdelnabi@microsoft.com} 
  } \\
}

\begin{document}
\maketitle

\begin{abstract}
\vspace{-0.2cm}

Large Language Models (LLMs) are increasingly utilized in autonomous decision-making, where they sample options from vast action spaces.
However, the heuristics that guide this sampling process remain under-explored.
We study this sampling behavior and show that this underlying heuristics resembles that of human decision-making: comprising a \textcolor{average}{descriptive} component (reflecting statistical norm) and a \textcolor{ideal}{prescriptive} component (implicit ideal encoded in the LLM) of a concept.
We show that this deviation of a \textcolor{sample}{sample} from the statistical norm towards a \textcolor{ideal}{prescriptive} component consistently appears in concepts across diverse real-world domains like public health, and economic trends.
To further illustrate the theory, we demonstrate that concept prototypes in LLMs are affected by \textcolor{ideal}{prescriptive} norms, similar to the concept of normality in humans.
Through case studies and comparison with human studies, we illustrate that in real-world applications, the shift of \textcolor{sample}{samples} toward an ideal value in LLMs' outputs can result in significantly biased decision-making, raising ethical concerns.
\end{abstract}

\vspace{-0.5cm}
\section{Introduction}
\label{Introduction}
\vspace{-0.2cm}

Decision making is a challenging task which often requires choosing an option from a vast set of possibilities~\citep{mattar2022planning,ross2023new}. 
In many real world cases deliberating on these innumerable options to decide on the action is computationally prohibitive, due to which, agents employ heuristics to sample their options~\cite{gigerenzer2011heuristic}.
For instance, humans (and animals) are shown to deliberate only on a few options that are selected based on a heuristics guided by possibility (how statistically likely an option is) and utility (the value associated to the option)~\cite{bear2020comes, mattar2018prioritized}.
While LLMs are often described as ‘System-1’ (Appendix~\ref{sec:glossary}) agents, characterised by their reliance on heuristics, the mechanism governing their response sampling remains under-explored.

\begin{figure*}[t]
    \centering
    \includegraphics[width=0.9\linewidth]{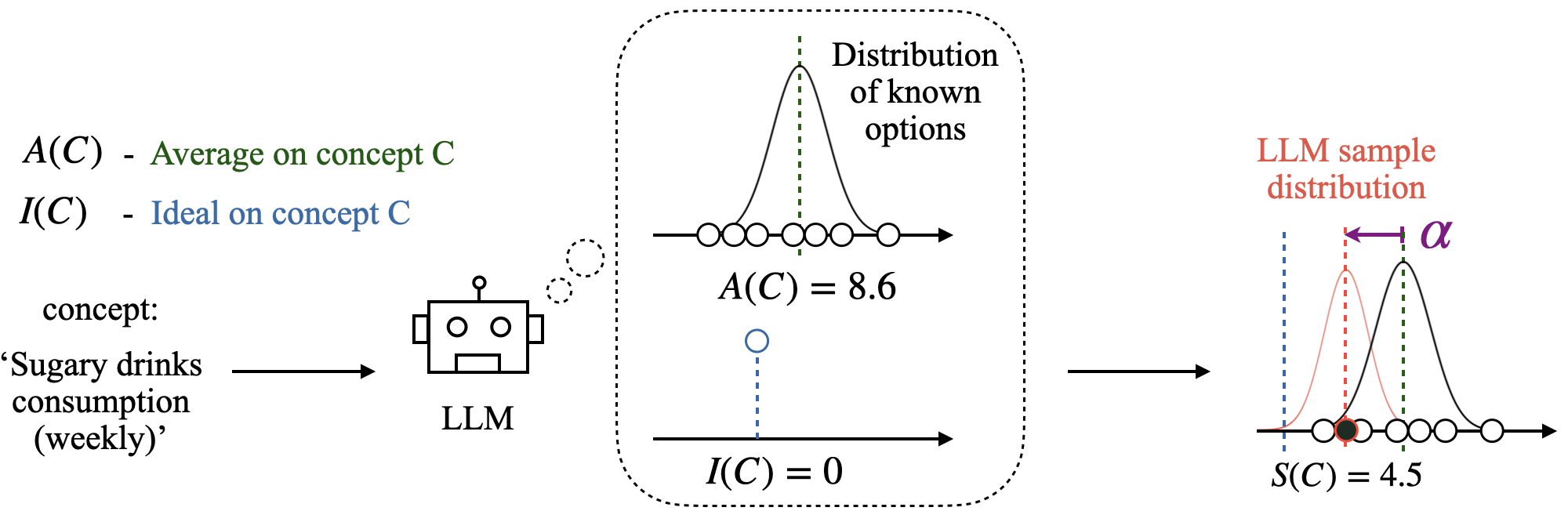}
    \vspace{-0.3cm}
    \caption{From left to right: when sampling on a concept, the LLM appears to account for the \textcolor{average}{statistical likelihood} (\textit{$A(C)$}) and \textcolor{ideal}{prescriptive norm} (\textit{$I(C)$}) of the concept. Consequently, the \textcolor{sample}{sample} distribution exhibits a shift (shown as \textit{$\alpha$}) away from the true distribution in the direction of the \textcolor{ideal}{ideal} (right most plot).}
    \vspace{-0.5cm}
    \label{fig:teaser}
\end{figure*}

We define response sampling as the process by which the LLM agent probabilistically selects outputs from a distribution of potential options (refer Appendix~\ref{sec:glossary} for all formal definitions).
We systematically study this sampling heuristics and show that they resemble that of human decision-making.
When an LLM samples from multiple possibilities of a concept, the sampling heuristics is driven by a \textcolor{average}{descriptive} component (the statistical norm of the concept) and a \textcolor{ideal}{prescriptive} component (a notion of an ideal of the concept).
A \textcolor{average}{descriptive} component represents what is statistically likely for a concept, reflecting the occurrence or probability of options.
A \textcolor{ideal}{prescriptive} component is an implicit standard of what is considered ideal, desirable, or a valued option of a concept. 
These norms on the concepts can be learned both in-context or in pre-training.

We design a critical experiment to isolate the effects of the proposed theory. We then show the effects of this heuristic appearing consistently across diverse real-world domains.
We perform extensive experiments covering different LLMs, evaluated concepts, and ablations to show the robustness of observations.
We present a medical case study where an LLM as an agent is used to assign a recovery time of patients to show potential practical concerns.
As illustrated in Figure~\ref{fig:teaser}, the proposed theory implies that when the LLM picks \textcolor{sample}{samples} for a concept, the \textcolor{sample}{sample} not only reflects the statistical regularities of the concept (\textcolor{average}{descriptive} norms) but also systematically incorporates an idealized version of the concept (\textcolor{ideal}{prescriptive} norms).
We show that these shifts may not align with human ideals, raising ethical concerns when LLMs are used for autonomous decision making.

Human decision-making is often guided by concept prototypicality~\cite{murphy2004big}, which incorporates both \textcolor{average}{descriptive} (statistically common) and \textcolor{ideal}{prescriptive} (value) components~\citep{barsalou1985ideals} - (e.g., while most teachers may exhibit certain average competence, the prototypical teacher is often imagined to teach well). We make initial investigation to show that LLMs' prototypicality has these two components and hypothesis its connection to sampling. In short, we make the following contributions:
\vspace{-0.2cm}
\begin{itemize}[leftmargin=*]
    \item We study the sampling mechanisms in LLMs through the lens of cognitive studies in humans. We show that the heuristics driving the sampling processes of both humans and LLMs converge on having a \textcolor{average}{descriptive} component and a \textcolor{ideal}{prescriptive} component. We construct an experimental setting to isolate the effect and empirically validate the proposed theory with many robustness checks and comparison with human studies.
    \vspace{-0.2cm}
    \item We evaluate \textcolor{sample}{samples} from a set of 500 existing concepts across 10 domains to verify the validity of the proposed theory. We find the results, on 15 language models covering different families and sizes, to be statistically significant. We show a case study inspired by real-world applications where this \textcolor{ideal}{prescriptive} component may lead to undesired outcomes.
    \vspace{-0.2cm}
    \item We demonstrate that LLMs' prototypical representations of concepts systematically incorporate \textcolor{ideal}{prescriptive} norms, showing initial evidence that their judgments of 'typical' examples are biased toward idealized versions, similar to human notion of prototypicality.  
    \end{itemize}

\section{Related Work}
\label{Related Work}

Earlier work that examined the mechanisms by which LLMs generate outputs suggests that they produce coherent text by probabilistically assembling language patterns without `genuine understanding'~\cite{bender2021dangers}. But, later investigations have demonstrated that LLMs can develop internal, structured representations of the environment~\cite{li2022emergent}. They even exhibit an understanding of semantic structures when trained on programming languages, indicating a capacity for meaningful text processing and generation~\cite{jin2023evidence}.
This has sparked interest within the community to explore the mechanisms governing output generation in LLMs through the lens of cognitive science and related disciplines.

Recent work indicates that LLM agents despite understanding the notion of probabilities struggle with probability sampling~\cite{gu2024llms}. They do not fully represent the statistics, i.e. poor at generating samples that align with expected probabilistic patterns. Our paper provides a systematic framework that explains the components in samples of LLMs. This can potentially explain the different biases shown by LLMs (more in Appendix \ref{realted_work_app}).

\textbf{Understanding LLMs as `System-1':} Reasoning has been broadly characterized as a two-step process involving quick `System-1' thinking and a more deliberate `System-2' reasoning~\citep{daniel2017thinking}.
Large Language Models (LLMs) have been conceptually likened to System-1 due to their heuristic-driven and non-deliberative output generation~\citep{yao2023tree}. In fact, recent studies show overlaps in errors made by LLMs and humans in System-1 reasoning tasks, indicating that both might rely on similar heuristics for rapid decision-making~\citep{dasgupta2022language}. We study the convergence of heuristics between LLMs and humans and propose a theory for LLM sampling.

Previous research mainly uses sampling for tasks like action generation and decision making rather than to explicitly study the sampling mechanisms of LLMs~\citep{hazra2023saycanpay, shah2023navigation, suri2023large}. Our work aims to fill this gap by investigating the heuristics driving LLMs' response sampling, which could provide a deeper understanding of their decision-making processes.

\section{Theory of LLM sampling}
\label{Method}

When faced with numerous possible actions, where deliberating on each option is computationally prohibitive, humans inherently resort to forming a finite consideration set of options using heuristics~\cite{phillips2019we}.
Cognitive studies characterize this heuristic-based filtering as `System-1' thinking: fast, automatic, and intuition-driven~\cite{daniel2017thinking, gigerenzer2011heuristic}.
Such heuristics effectively reduce the cognitive load on deliberative processes (`System-2'), by selecting a manageable subset of options for further deliberation.
In humans, these heuristics are guided primarily by two factors: the statistical likelihood of options and their perceived value~\cite{bear2020comes}.

In LLMs, reasoning mechanisms such as CoT~\cite{wei2022chain} and other explicit reasoning models like GPT-o3~\cite{openai2024reasoning} and Deepseek-r1~\cite{guo2025deepseek} are likened to explicit deliberation (`System-2'), while the default mechanism is likened to heuristics driven `System-1'~\cite{li2025system}.
Hence, understanding the heuristics driving their sampling is key to explaining their performance.
We examine the sampling mechanisms of LLMs in the light of this human cognitive theory and propose a theory for LLM sampling:

\begin{tcolorbox}[colframe=black, arc=2mm, halign=center]
Sampling of an LLM is driven by \textcolor{average}{descriptive} component (the statistical norm of the concept) and a \textcolor{ideal}{prescriptive} component (a notion of an ideal of the concept).
\end{tcolorbox}

This implies that, when an LLM samples from multiple possibilities of a concept, the heuristics is driven by the statistical norm of the concept and a notion of an ideal of the concept. Here, sampling is defined as the process by which the model probabilistically selects outputs from a distribution of potential responses. We refer the reader to the glossary (Appendix \ref{sec:glossary}) for the detailed definitions of all terms. 

In humans, these two components of thought is hypothesized to originate from them being goal-driven agents and engaging in value maximization~\citep{bear2017normality}. On the other hand, the underlying auto-regressive mechanism of LLMs is not goal-driven, it is non-trivial how the \textcolor{sample}{sample} has a \textcolor{ideal}{prescriptive} component. The experimental methodology of this work is exactly following established principles in uncovering heuristics of humans in the cognitive science literature ~\citep{bear2020comes,phillips2019we}.

\subsection{Sampling in relation to a novel concept}
\label{value_bias_in_context}

The proposed theory calls for a rigorous validation and for this we use an established framework used in humans~\cite{bear2020comes} and further scale it for more evidence. This well-founded setup is a critical experiment providing compelling evidence in support of our proposed theory. In this setting, we introduce a novel concept \textit{$C$} to eliminate potential confounding effects associated with using pre-existing concepts embedded in the LLM. We present the LLM with the exact same prompt but varying \textcolor{average}{descriptive} and \textcolor{ideal}{prescriptive} components for the concept \textit{$C$}. We evaluate the output \textcolor{sample}{samples} to show the effect of the two varying components (\textcolor{ideal}{prescriptive} and \textcolor{average}{descriptive}) on sampling.

To establish a statistical baseline for concept \textit{$C$}, we use numbers from a Gaussian distribution  with mean \textit{$C_{\mu}$} (and known variance).
The LLM is provided with \textit{$N$} samples from this distribution representing possible options associated with concept \textit{$C$}.
To ensure the reliability of the baseline, \textit{$N$} is chosen to be sufficiently large that the mean of the input samples closely approximates \textit{$C_{\mu}$}.
Following this, to establish a \textcolor{ideal}{prescriptive} norm \textit{$C_v$} on the concept \textit{$C$}, we associate each of the $N$ options with a \textcolor{ideal}{prescriptive} component, represented by a grade. 

We run the experiment with the following setting for \textit{$C_v$}: a higher value being \textcolor{ideal}{ideal}, a lower value being \textcolor{ideal}{ideal}, and a control experiment having no explicit ideal direction. Based on the input (the \textit{$N$} samples along with the corresponding grades), we prompt the LLM to provide a \textcolor{sample}{sample} for the concept \textit{$C$}. We denote this \textcolor{sample}{sample} reported by LLM on concept \textit{$C$} as \textit{$S(C)$}.
By systematically changing $C_{\mu}$ and $C_{v}$ and keeping the rest of the prompt same, we study the corresponding change in \textcolor{sample}{samples} \textit{$S(C)$}.

For each \textit{$C_{\mu}$} and \textit{$C_v$}, in independent contexts (i.e., prompts), we repeat the procedure \textit{$M$} times to obtain a \textcolor{sample}{sample} distribution.
We keep the value of \textit{$M$} the same as \textit{$N$} in all variants of the experiments to compute statistical significance of the shift in input and \textcolor{sample}{sample} distribution.
If the \textcolor{sample}{sample} is driven solely by the \textcolor{average}{descriptive} norm (statistics of the input samples), the distribution of \textcolor{sample}{samples} \textit{$S(C)$} is expected to be statistically similar to the input distribution.

The difference between input samples and the \textcolor{sample}{samples} reported by LLM might occur also due to the error in approximating the statistics of the input samples, i.e the LLMs' inability to `understand' the statistics of the distribution. To exclude this possibility, we instruct the LLM to report the \textcolor{average}{average} of the distribution. We denote the reported \textcolor{average}{average} by \textit{$A(C)$}. Across all experiments, we observe that \textit{$C_\mu$} $\approx$ \textit{$A(C)$}, indicating that the LLM reliably approximates the statistics of the input distribution.

We apply the Mann-Whitney U test to compare the distribution of \textcolor{sample}{samples} \textit{$S(C)$} with (a) input distribution and (b) distribution of reported averages \textit{$A(C)$}. For each concept, \textit{$C$}, we calculate the Mann-Whitney U statistic and the corresponding $p$-value. If $p<0.05$, there is a significant difference between the evaluated distributions. We vary the direction of \textit{$C_v$} and demonstrate that the change in \textcolor{sample}{samples'} mean (mean of \textit{$S(C)$}) corresponds to the change in \textit{$C_v$}. As a sanity check we do this experiment without any grades to show that the LLM can indeed approximate the input distribution. Hence the deviation in the direction of \textcolor{ideal}{ideal} conclusively demonstrate that the observed deviation in sampling is indeed a heuristics of the LLM and not coming from an incapability to approximate distribution.

\subsection{Sampling in relation to existing concepts}
\label{implicit_value_bias}
In this section, we investigate the validity of the theory beyond the constructed setting on five hundred existing concepts in the LLM across ten domains. For an existing concept, the statistics of possible options and associated values are already embedded in the LLM and not known to us. That is \textit{$C_{\mu}$} and \textit{$C_v$} associated to the concept $C$ is not known.

Similar to the previous setting, for a concept \textit{$C$}, we evaluate the statistical difference between \textit{$A(C)$} and \textit{$S(C)$} to show the validity of the proposed theory. We use \textit{$I(C)$}, the self reported \textcolor{ideal}{ideal} value to get the direction of \textit{$C_v$}. We use a binomial test to determine whether the \textcolor{sample}{sample} \textit{$S(C)$} falls on the \textcolor{ideal}{ideal} side of the \textcolor{average}{average} \textit{$A(C)$} or its non-ideal side. The latter can also be understood as the \textcolor{sample}{sample} falling on the \textcolor{average}{average} side of \textcolor{ideal}{ideal}.

\begin{figure*}
    \centering
    \includegraphics[width=0.9\linewidth]{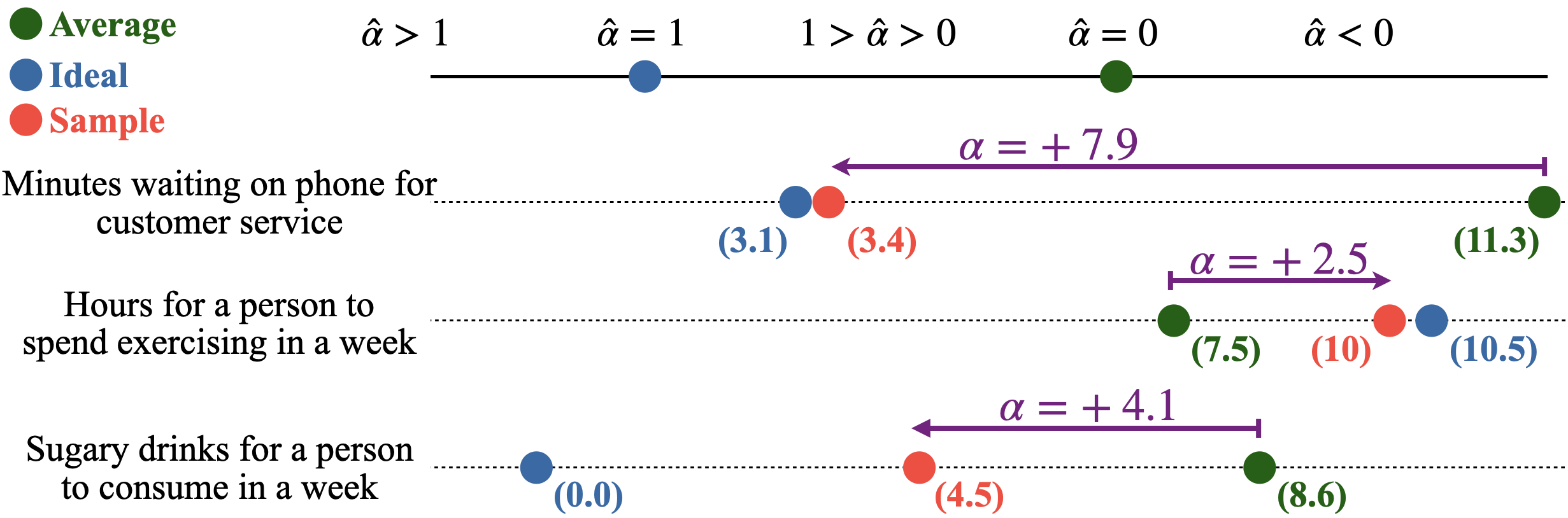}
    \caption{The figure shows the \textcolor{average}{average}, \textcolor{ideal}{ideal}, and \textcolor{sample}{sample} values reported by the LLM for three different concepts. Positive \textit{$\alpha$} shows the deviation in the direction of the \textcolor{ideal}{ideal}.} 
    \label{fig:implicit}
\end{figure*}

Examples of this framework are shown in Figure \ref{fig:implicit}. Consider the number of concepts for which \textcolor{sample}{sample} falls on the \textcolor{ideal}{ideal} side of the \textcolor{average}{average} is \textit{$n$} and the total number of concepts evaluated is  \textit{$n_{total}$}. The Binomial test is used to determine if \textit{$n$} is significantly different from what would be expected by chance, assuming a null hypothesis where the probability \textit{$p$} of a \textcolor{sample}{sample} being on the \textcolor{ideal}{ideal} side is 0.5. The $p$-value obtained from the binomial test is used to assess significance. $p<0.05$ shows presence of \textcolor{ideal}{prescriptive} norm across concepts.

\textbf{Drift from the statistical norm: }In most applications, one might expect the LLM to sample options based on their statistical likelihood. We use a variable \textit{$\alpha$} to quantify the degree to which the \textcolor{sample}{sample} deviates away from the statistical norm. We define \textit{$\alpha$} such that, the value of \textit{$\alpha$} is positive when the proposed theory holds. That is, \textit{$\alpha$} is measured to be positive when \textit{$S(C)$} deviates from the \textit{$A(C)$} in the direction of \textit{$C_v$} or \textit{$I(C)$}. \textit{$\alpha$} is shown in both the figures (\ref{fig:implicit} and \ref{fig:teaser}). Formally, for each \textcolor{sample}{sample} \textit{$S(C)$} of a concept \textit{$C$}, \textit{$\alpha$} is computed as 

\begin{equation}
    \alpha =  (\textit{A(C)} - \textit{S(C)} ) \times \textit{sign}(\textit{A(C)} - \textit{I(C)}) 
\end{equation}

We also compute $\hat{\alpha}$: a normalized scale such that \textit{$A(C)$} is at the origin and \textit{$I(C)$} is at unit distance from the origin. We compute $\hat{\alpha}$ as $\alpha/ \lvert \textit{A(C)} - \textit{I(C)}\rvert$. $\hat{\alpha}$ enables comparison across concepts with less dependency on the scale of values. It also allows comparison with observations obtained in the experiments with human subjects. 

The deviation metric $\alpha$ is defined as a directional measure quantifying the shift from the statistical norm \textit{$A(C)$} toward the \textcolor{ideal}{prescriptive} norm \textit{$I(C)$}. However, when the average and ideal values are equal (i.e., \textit{$A(C) = I(C)$}), the directional term \textit{sign($A(C) - I(C)$)} becomes zero, leading to an undefined $\alpha$ and $\hat{\alpha}$. These degenerate cases are excluded from the $\alpha$ based analysis, as no directional deviation can be computed. Such analysis is meaningful when there is a difference between statistical and \textcolor{ideal}{prescriptive} components. 

\textbf{Comparing with human studies:} The setting described in Sections \ref{value_bias_in_context} and ~\ref{implicit_value_bias} is inspired by similar evaluation in humans~\citep{bear2020comes,phillips2019we,bear2017normality}. We scale the experiments to show higher statistical significance. In appendix, we show replication of the exact setting of human studies (Table \ref{tab:bear_paper}) on LLM (Table \ref{tab:implicit value bias}) to make a direct one to one comparison using the respective \textit{$\alpha$}s (Figure \ref{fig:alpha_plot_full} (left)). \textbf{The conclusion follows that the heuristics of sampling converges in LLMs and humans but the degree of deviation of sample towards the \textcolor{ideal}{prescriptive} norm does not align}. This causes deviation of \textcolor{sample}{sample} from the \textcolor{average}{statistical likelihood} in unforeseen degrees, an interesting direction for future research in fairness and alignment.

\section{Experiments and Results}

In this section, we present two key experiments and a case study. First, we present a constrained setting to test the validity of the proposed theorem following the method in Section \ref{value_bias_in_context}. Second, we evaluate the presence of \textcolor{ideal}{prescriptive} and \textcolor{average}{descriptive} components in sampling for concepts learned in training following Section \ref{implicit_value_bias}. Our results show significant evidence for the proposed theory. We test on the instruction-tuned models of GPT-4~\citep{achiam2023gpt}, GPT-3.5-Turbo~\citep{brown2020language}, Claude~\citep{anthropic2024claude}, Mixtral-8x7B~\citep{jiang2024mixtral}, Mistral-7B~\citep{jiang2023mistral}, and both pretrained and instruction tuned models from the family of Llama-2 and 3 models~\citep{touvron2023llama}~\ref{sec:compute_resources}. Unless mentioned otherwise, we report results for GPT-4 in the main text and the results for other models in the Appendix. The complete text used in the prompts for each experiment is given in the Appendix~\ref{sec:exp2_prompts}, \ref{sec:exp1_prompts}, \ref{full_cat} and \ref{sec:exp3_prompts} respectively. 

\subsection{Sampling in relation to a novel concept}
\label{exp:glubbing}

Following Section~\ref{value_bias_in_context}, we empirically validate the proposed theory by constructing a constrained setting around a novel, fictional concept: ``glubbing''. We also, consider multiple such random fictional concepts defined in different terms (Appendix~\ref{sec:exp2_glubbing_variants_gpt4}). 

We systematically vary $C_{v}$ and $C_{\mu}$ to study the effect on the distribution of \textcolor{sample}{samples} \textit{$S(C)$}. The rest of the prompt is kept similar to isolate the influence of \textcolor{average}{descriptive} and \textcolor{ideal}{prescriptive} components in the LLM’s sampling process so that there is no interference from prior knowledge and prompt artifacts. Importantly, the results are designed to reflect the heuristics of sampling of the LLM, independent of the prompt design or specific experimental conditions. The prompt contains (a) statistical norm defined by a hundred samples from a distribution corresponding to hours spent ``glubbing'' and (b) $C_{v_i}$ the ideality associated with each sample $i$. $C_{v_i}$ is given by a grade on a scale of \textit{A+} to \textit{D-}.

In this first run, ``glubbing'' hours of people is sampled from a Gaussian of mean 45 and a standard deviation of 15. We repeat the experiment with a bi-modal Gaussian distribution with modes at 35 and 65 and a standard deviation of 5. The implementation and analysis of the two experiments are the same. 

We evaluate the value system \textit{$C_v$} in three levels of valence: (a) positive, (b) negative, and (c) neutral (control experiment). For the positive \textit{$C_v$}, the grades are assigned such that the higher hours of ``glubbing'' get a better grade (best being \textit{A+}), and for the negative value system, the grades are assigned such that the lower hours of ``glubbing'' get a better grade (on the same scale). A sample positive prompt is given below: 

\begin{tcolorbox}[title={Prompt},colframe=black!30!white]
There is a hobby called glubbing. Here are the glubbing hours of people and a grade associated, A+ being the highest grade and D- being the lowest grade: 43:C, 35:C-, 63:B+, \ldots,  35:C-. <sample prompt>.
\end{tcolorbox}

The `\ldots' corresponds to the rest of the values and grades (the prompt has a hundred samples and corresponding grades). The full prompt set is given in Appendix \ref{sec:exp2_prompts}. The vanilla <sample prompt> is: `Based on this, pick a sample number of glubbing hours'. Different sample prompts gives similar results as shown in~\ref{sec:exp2_glubbing_variants_gpt4}. 

A shift between input distribution and \textcolor{sample}{sample} distribution can be explained as the error of LLM in approximating the statistics of the input distribution. To exclude this alternative explanation, we compute the significance in the shift of generated \textcolor{sample}{samples} (\textit{$S(C)$}) from the \textcolor{average}{average} reported by the LLM (\textit{$A(C)$}). In the neutral control experiment, we assign the mean $C_{\mu}$ with the highest grade and lower grades for increasing distance from the mean. We run the experiment for positive, negative, and control settings a hundred times each.

\textbf{Results:} Table~\ref{tab:glubbing_results} shows the result for the mean of the hundred runs for the uni-modal and bi-modal input distributions, each with three different \textit{$C_v$}. Firstly, across the six settings, \textit{$A(C)$} approximately coincide with the true distribution average (\textit{$C_{\mu} = 45$}). For a neutral \textcolor{ideal}{prescriptive} norm (also for no \textcolor{ideal}{prescriptive} norm as shown later), \textit{$S(C) \approx A(C) \approx C_\mu$} and the input distribution and \textit{$S(C)$} do not differ significantly, $\textit{p} = 0.52$. Given the non-significant difference ($\textit{p} = 0.52$), the result is consistent with the hypothesis that sampling reflects the input distribution’s statistical properties when no ``ideal'' is specified.


When \textit{$C_v$} is positive, the mean of \textcolor{sample}{samples} is higher than the mean of the LLM-generated \textcolor{average}{average}. Also for negative \textit{$C_v$}, the mean of \textcolor{sample}{samples} is lower than the mean of the LLM-generated \textcolor{average}{average}. For instance, in the uni-modal scenario, the mean \textit{$S(C)$} for negative \textit{$C_{v}$} is 36.5, and positive \textit{$C_v$} is 46.7. 

When \textit{$C_v$} is positive, the distribution of \textit{$S(C)$} and distribution of \textit{$A(C)$} are significantly different, with \textit{p}$ = .003$, and for a negative \textit{$C_v$}, \textit{p}$ < .001$. 
\textbf{This shows that the \textcolor{sample}{sample} is not solely driven by the \textcolor{average}{statistics} of the input distribution, but also the \textcolor{ideal}{prescriptive} norm of the concept.}

\begin{figure}[t]
    \centering
    \includegraphics[width=0.8\columnwidth]{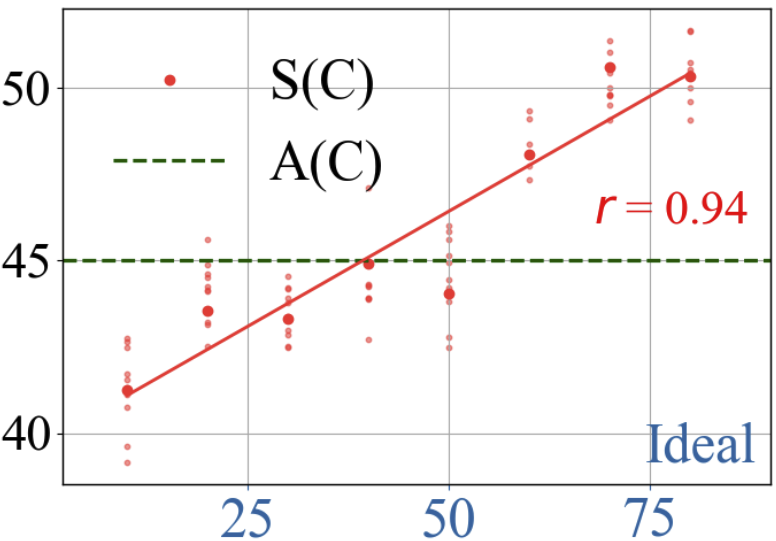}
    \caption{Variation of mean of \textit{$S(C)$} with changing \textcolor{ideal}{prescriptive} value. X-axis shows the different \textcolor{ideal}{prescriptive} values and Y-axis shows the \textcolor{sample}{sample} value. \textcolor{sample}{Sample} is directly proportional to \textcolor{ideal}{prescriptive} value. Here \textit{$C_{\mu}= 45,$} approximately equal to \textit{$A_{C}$}. }
    \label{fig:result}
\end{figure}

\textbf{Robustness of the experiment: }

We vary the mean \textit{$C_{\mu}$} of the input distribution to show the reliability of the conclusion in Appendix~\ref{sec:varion_with_mu}. We show that for a range of \textit{$C_{\mu}$}, \textit{$A(C)$} $\approx$ \textit{$C_{\mu}$} and for each of this \textit{$C_{\mu}$}, \textit{$S(C)$} consistently shifts away from \textit{$A(C)$} in the direction of \textit{$C_v$}. We also repeat this experiment with different newly introduced fictional scenarios (different tokens other than ``glubbing'' used to define the new concept) and also introduced them as different ideas (not just as a hobby, details in Appendix~\ref{sec:exp2_glubbing_variants_gpt4}). As an additional control, we repeat this experiment by assigning no grades and random grades to the input samples. We found no significant shift in the distribution of input samples and \textit{$S(C)$}, \textit{p}$ = 0.51$ and \textit{p}$ = 0.52$ respectively. 

Note that, to ensure the observation is not merely an artifact of the prompt, we use the same prompt in all cases, varying only \textit{$C_v$} across the three runs. To further validate robustness of observation to the prompt, we use different <sample prompt> in Appendix~\ref{sec:exp2_glubbing_robustness_gpt4}. Results show that our conclusion holds for these variations. In Appendix~\ref{sec:exp2_glubbing_robustness_gpt4}, \ref{robustness_prompt} we also show strong results that even explicit debasing prompt fails to undo the \textcolor{ideal}{prescriptive} component.

We scale this experiment by varying \textit{$C_{\mu}$} in the range of $45$ to $845$ in intervals of hundreds. For each \textit{$C_{\mu}$} we give eight different grading scheme: varying the number which gets the best grade in the intervals of ten. The grade reduces with distance on either side of the number with best grade (like a tent function). Each of the combination of \textit{$C_{\mu}$} and peak ideal is run hundred times and the mean deviation of sample is reported. An example plot of \textit{$C_{\mu}$=45} and $8$ different peak ideal values is in Figure{~\ref{fig:result}}. Rest of the plots are in Figure~\ref{fig:beuty}. We see the pattern of \textcolor{sample}{sample} consistently shifting from the \textcolor{average}{descriptive} component towards \textcolor{ideal}{prescriptive} component across the different runs.

\begin{table}[ht]
  \centering
  \begin{tabular}{l c c c c}
    \toprule
    & \multicolumn{2}{c}{uni-modal} & \multicolumn{2}{c}{bi-modal} \\
    \cmidrule(lr){2-3} \cmidrule(lr){4-5}
    & \textit{$A(C)$} & \textit{$S(C)$} & \textit{$A(C)$} & \textit{$S(C)$} \\
    \midrule
    C\_v: +ve     & 44.94 & 46.72 & 44.97 & 47.43 \\
    C\_v: -ve     & 44.99 & 36.50 & 45.03 & 41.26 \\
    C\_v: control & 45.01 & 44.95 & 44.94 & 44.95 \\
    \bottomrule
  \end{tabular}
      \caption{The table shows the change in mean of \textcolor{sample}{samples} (mean of \textit{$S(C)$}) and the mean of reported \textcolor{average}{average} (mean of \textit{$A(C)$}). For these experiments \textit{$C_{\mu} =45$}, the result for other \textit{$C_{\mu}$} is given in Appendix~\ref{sec:varion_with_mu}.}
  \label{tab:glubbing_results}
\end{table}

We observe statistically significant results for most other evaluated LLMs. Results for GPT-4 (with temperature set to zero), GPT-3.5-Turbo, Claude, Mixtral-8x7B, Mistral-7B, and Llama models are in Appendix \ref{sec:glubbing_other_LLM}.  For example, Claude-Opus, with a negative and positive \textit{$C_v$}, \textit{$S(C)$} is statistically significant from \textit{$A(C)$} with \textit{p}$<.001$.

\subsection{Sampling in relation to existing concepts}
\label{implicit}

In this experiment, the statistics \textit{$C_{\mu}$} and value system \textit{$C_v$} for a concept \textit{$C$} are implicit in the LLM and unknown to us. We empirically evaluate the proposed theory on \textbf{500} different concepts (\textit{$C$}) spanning \textbf{10} different domains. The full list of concepts are in the Appendix~\ref{full_cat}. For each concept, we first ask the model to report its notion of (a) the \textcolor{average}{average} (\textit{$A(C)$}) and (b) the \textcolor{ideal}{ideal} (\textit{$I(C)$}) for a given concept \textit{$C$}. We then give a sample prompt for concept \textit{$C$} to get (c) \textcolor{sample}{sample} (\textit{$S(C)$}). These prompts are given in independent contexts. To get these values, we use a prompt similar to the questions used in human studies~\citep{bear2020comes}. For example, to get the \textcolor{average}{average}, \textcolor{ideal}{ideal}, and the \textcolor{sample}{sample}  on the concept of `TV watching hours of people', we use the following prompts:

\begin{tcolorbox}[title={Prompt for Implicit Prescriptive Norms},colframe=black!30!white, width=\linewidth, boxrule=0.5mm, colback=white]
\begin{tabular}{@{}m{0.5cm} m{7.5cm}@{}}
\textcolor{average}{$P_a$}: & \parbox[t]{5.5cm}{What is the average number of hours of TV a person watches in a day?}\\
\textcolor{ideal}{$P_i$}: & \parbox[t]{5.5cm}{What is the ideal number of hours of TV for a person to watch in a day?}\\
\textcolor{sample}{$P_s$}:  & \parbox[t]{5.5cm}{What is the number of hours of TV a person watches in a day?}
\end{tabular}
\end{tcolorbox}

\begin{table}[h!]
\small
\centering
\begin{tabular}{l l l}
\toprule
\textbf{Model Name} & \textbf{Significance} & \textbf{Fraction} \\
\midrule

Llama-2-7b  & 6.837e-02 & 0.539 \\
Llama-2-7b instruct  & 3.874e-06 & 0.607 \\
Llama-2-13b  & 3.952e-06 & 0.613 \\
Llama-2-13b-chat  & 3.023e-10 & 0.642 \\
Llama-2-70b  & 4.496e-07 & 0.622 \\ 
Llama-2-70b-chat  & 1.583e-16 & 0.688 \\
Llama-3-8b  & 1.109e-05 & 0.608 \\
Llama-3-8b-Instruct  & 9.277e-22 & 0.716 \\
Llama-3-70b  & 3.041e-21 & 0.726 \\
Llama-3-70b-Instruct  & 5.382e-35 & 0.777 \\
Claude  & 1.582e-16 & 0.688 \\
Mixtral-8x7B   & 9.289e-22 & 0.716 \\
Mistral-7B   & 1.114e-05 & 0.608 \\
GPT-4 & 5.506e-15 & 0.680 \\
\bottomrule
\end{tabular}
\caption{Model comparison across LLMs showing influence of the \textcolor{ideal}{prescriptive} component in existing concepts.The fraction indicates the proportion of concepts within each domain for which the LLM’s sampled value deviates from the average in the direction of the ideal. The table shows a larger influence of \textcolor{ideal}{prescriptive} norms for larger model sizes and higher for RLHF compared to pretrained-only models.}
\label{tab:implicit_value_bias2}
\end{table}

\textbf{Results:} In GPT-4, for each concept, we run the three prompts ten times with a temperature of 0.8 and report the average in Table~\ref{tab:implicit_value_bias2}. Prompts failed for 10 concepts and the value of \textit{$A(C)$} and \textit{$I(C)$} were the same for 46 concepts. For the rest of the concepts, we observe that 304/444 samples fall on the ideal side of average (positive \textit{$\alpha$}). This gives a statistical significance of $5.06 \times 10 ^ {-15}$, a \textbf{very high statistical significance}, reducing the likelihood of the result being due to chance. The result gives strong evidence to the proposed theory. 

Except for the Llama-2-7b base, all the other LLMs show a statistically significant deviation towards the \textcolor{ideal}{prescriptive} norm and even this model is only marginally insignificant. We also make the following observations: 
\begin{itemize}[leftmargin=*]
    \item The influence of \textcolor{ideal}{prescriptive} norms seems to get larger as the model size increases.
    \item \textcolor{ideal}{Prescriptive} norm seems to stem from pretraining rather than RLHF, though RLHF exacerbates it.
\end{itemize}

Our results suggest that the significance of the observation tends to increase with model size/capability.
Such an `inverse scaling law'~\citep{mckenzie2023inverse} should be taken into account in scenarios like the case study given below.

\paragraph{Case study for medical recovery time:} Deviation of a \textcolor{sample}{sample} towards the \textcolor{ideal}{prescriptive} norm can help explain some biases of LLMs. To illustrate this, we present a case study based on a real-world scenario. The LLM agent is assigned the role of a doctor and asked to take a decision on the discharge time of a patient based on a list of symptoms. Here the action space is the positive rational numbers (number of weeks). Once the LLM gives a recovery time we also get self reported \textcolor{average}{average} and \textcolor{ideal}{ideal} recovery time from the LLM. The term self-reported average refers explicitly to the average values directly provided by the LLM itself when prompted to report average.

The setup is similar to Experiment~\ref{implicit}, but we prompt the LLM to be a doctor and give output (in weeks) based on a given list of four symptoms. We find that the LLM significantly deviates from statistical norm recovery time towards a notion of an \textcolor{ideal}{ideal} when one might assume and, in fact in this example, \emph{require} that the LLM is using only the statistical norm. Out of the 35 symptom batches (each of four symptoms), the \textcolor{sample}{sample} falls on the \textcolor{ideal}{ideal} side of \textcolor{average}{average} 26 times-a statistically significant shift (binomial $p$ = 0.003). 

The \textcolor{ideal}{ideal} value given by the LLM, is lower than the \textcolor{average}{average} value in 30 of the 35 symptoms. This implies that the \textcolor{sample}{sample} is often pulled below the \textcolor{average}{average}. This finding indicates that LLMs' decision-making regarding patient recovery times is compromised by a \textcolor{ideal}{prescriptive} component, which has significant implications for clinical decision-making, resource allocation in hospitals, and potential risks to patient safety. The full list of the symptoms and the exact prompts used is given in the Appendix \ref{sec:exp1_case_study}.

\section{Prescriptive component in concept prototypes}
\label{prototype}

One of the basic characteristics of System-1 is that it represents concepts with prototypical examples~\cite{daniel2017thinking}. In humans, though a prototype is often understood as the most typical/representative member of a concept~\citep{murphy2004big}, they are found to embody both \textcolor{average}{statistical} regularities and goal-oriented \textcolor{ideal}{ideals} within a concept~\citep{barsalou1985ideals}.

For instance, a `Robin' might be considered a prototype of the concept `Bird', as it shares many common features with most birds with high occurrence (statistics), and has the ability to fly (a value expected of birds), making it a prototypical example of the concept `bird'~\citep{smith1981categories}. 
For this reason, penguin-a flightless bird, is less prototypical bird than `Robin'. Prototypicality defines normality of a concept that drive the sampling (Appendix ~\ref{mot_prts}). 

Unlike humans, it is not clear whether LLMs rely on concept prototypes for sampling. But since the sampling heuristics of LLMs converge with humans, it is interesting to investigate concept prototypicality in LLMs. We do not claim that LLMs output is prototype driven, but make an initial exploration in this direction using the exact setting as in~\cite{bear2017normality}.

We use eight concepts, and for each concept \textit{$C$}, six different exemplars. Exemplars are short descriptions of items of a concept. For instance, for the concept of `High-school teacher', the first exemplar is as follows: `A 30-year-old woman who basically knows the material she is teaching but is relatively uninspiring, boring to listen to, and not particularly fond of her job'.

Similar to experiment protocol in~\citet{bear2017normality}, LLMs rate each exemplars on three dimensions: \textcolor{average}{average}, \textcolor{ideal}{ideal}, and the prototypicality of the exemplar. Prototypicality score is derived by averaging three entities, which measure the degree to which the given prototype is a ``good example'', ``paradigmatic example'', or ``prototypical example''~\cite{bear2017normality}. The LLM is asked to rate on a 7-point scale ranging from not at all average/ideal/`good example', which has a score of 0, to completely average/ideal/`good example', with a score 7. The full set of concepts and exemplars are in Appendix \ref{sec:exp3_prompts}.

As in the previous section \ref{implicit_value_bias}, we check whether the prototypicality rating of the concepts falls on the \textcolor{ideal}{ideal} side of the \textcolor{average}{average}.
To test significance, we do a binomial test across concepts to check if LLMs conception of prototypes has a \textcolor{ideal}{perspective} component. The evaluation is similar to the previous section.

\begin{table}[ht]
\small
\begin{tabular}{l|SSS[table-format=2.2]}
\toprule
Concept & \textcolor{average}{Average} & \textcolor{ideal}{Ideal} & \textcolor{sample}{Prototype} \\
\midrule
High-school teacher &{2.75} & {3.66} & {3.86} \\
Dog &{3.08} & {3.83} & {3.86} \\
Salad &{4.5} & {4.5} & {5.44} \\
Grandmother &{4.16} & {4.66} & {4.75} \\
Hospital &{2.91} & {3.5} & {3.55} \\
Stereo speakers &{2.92} & {4.16} & {3.61} \\
Vacation &{3.08} & {4.75} & {4.63} \\
Car &{2.58} & {4.083} & {4.11} \\
\bottomrule
\end{tabular}
\caption{Concepts and scores averaged across exemplars showing how the prototypical score doesn't coincide with just the \textcolor{average}{average} but also has an \textcolor{ideal}{ideal} component.}
\label{tab:matrix}
\end{table}

We run this experiment ten times on GPT-4 with a temperature of 0.8 and report the average results. The average scores from the three prototypicality assessments (``good'', ``paradigmatic'', and ``prototypical'' example) demonstrate satisfactory internal consistency, with a Cronbach's alpha of 0.96. Consequently, these scores were combined to form a single, comprehensive prototypicality rating. The aggregate results for each concept, averaged across exemplars, are given in Table~\ref{tab:matrix}. The results show a significant effect of a \textcolor{ideal}{prescriptive} component with 39 out of 46 falling on the \textcolor{ideal}{ideal} side of the \textcolor{average}{average} (binomial $p$ $<$ 0.001).

Evaluating across different LLMs, we obtain the following results: Llama-3-7b (binomial $p$ = 0.003), Mixtral-8x7B (binomial $p$ = 0.05), GPT3.5-turbo (binomial $p$ < 0.001),  Claude (binomial $p$ < 0.001), Mistral (binomial $p$ = 0.0019), indicating the effect of \textcolor{ideal}{prescriptive} norms in prototypes of concepts. The complete set of results for every exemplar is given in Appendix \ref{sec:exp3_results}. This experiment is an initial exploration, finding that LLMs' concept of prototypes is influenced not only by  \textcolor{average}{statistical} averages but also by an underlying  \textcolor{ideal}{prescriptive} norm. These findings suggest that the LLM's judgment of what constitutes a typical or prototypical example is systematically biased toward idealized representations calling for further investigations in this direction. 

\section{Comparison with human studies}
\label{exp:human}

The critical experiment presented in Section~\ref{exp:glubbing} is inspired by prior work with humans~\cite{bear2020comes}. In Appendix~\ref{sec:flubbing}, we present the results of the study conducted on human subjects. We replicate the exact setting using an LLM with a human-like prompts (Appendix \ref{sec:human}) and report the results in a similar visualization to facilitate a direct comparison. Our results show that, given a new fictional concept with \textcolor{ideal}{prescriptive} and \textcolor{average}{descriptive} statistics, humans and LLMs capture these norms and use them to sample options. Morever, we also note that LLMs, like humans seem to have an asymmetric treatment of gains and losses. The undersampling of negative value scenarios is observed to be more than oversampling of the positive value scenarios, potentially pointing to a shared optimism bias. This asymmetry, where both systems avoid negatives more strongly than they pursue positives, can be explored in future work.

Furthermore, we also create exact setting for experiment~\ref{implicit_value_bias} and compare the LLM and human outputs in Appendix~\ref{sec:exp1_humans} for known concepts. Here we use the same forty concepts used in humans studies to compare the results of LLM and humans. The comparison shows consistency of \textcolor{ideal}{prescriptive} influences across both human cognitive processes and LLM sampling. While both systems exhibit \textcolor{ideal}{prescriptive} components in sampling, a key divergence emerges in their treatment of ideals. Humans consistently conceptualize ideal as modest improvements over statistics (e.g., ideal `number of sugary drinks per week' is $2.41$ while the average is $9.17$), whereas LLMs frequently default to absolute and stricter ideals (e.g., $0$ for sugary drinks and $18$ other concepts) indicating moral absolutism which could be a topic of future investigation.

Finally, the investigation on prototypes presented in Section~\ref{prototype} follows the same prompt as human studies. Hence we do not need an explicit recreation of the human prompt for this experiment. This shows initial evidence that concept prototypicality scoring in LLMs are driven by the same components as in humans. Interestingly, scatter plot of $\hat{\alpha}$ for LLMs and humans for prototypicality~\ref{fig:alpha_plot_full} (right) show that, the amplitude of the influence of \textcolor{ideal}{prescriptive} component in sampling also correlate with that of humans ($\hat{\alpha}$ Pearson correlation of 0.33).

As a final note, our experiments probe systematic patterns in LLM outputs, revealing biases and decision tendencies analogous to human behavioral studies. We do not presuppose that these patterns emerge from human-like reasoning mechanisms.

\section{Conclusion}
\label{Conclusion}

In this paper, we set out to better understand the heuristics governing possibility sampling process of LLMs. Based on human cognitive studies, we propose a theory that explains the sampling heuristics to be part \textcolor{average}{descriptive} and part \textcolor{ideal}{prescriptive}. However, the exact \textcolor{ideal}{prescriptive} component might not be aligned with humans. As LLMs continue to be integrated into real-world applications, understanding their decision-making heuristics becomes increasingly important. Our results provide a foundational framework for evaluating how LLMs balance \textcolor{average}{statistically} probable outcomes with norms of \textcolor{ideal}{ideality}, raising interesting questions about their underlying representations. As a final remark, we would like to emphasize that we do not intend to contribute to ``humanizing'' AI/ML/LLMs in the way we use terminology or models. Instead, our contribution is intended to draw parallels in behaviour and perform evaluations, as our findings can have an impact on downstream tasks.

\section{Acknowledgements}

This work was partially funded by ELSA – European Lighthouse on Secure and Safe AI funded by the European Union under grant agreement No.101070617. Views and opinions expressed are, however, those of the authors only and do not necessarily reflect those of the European Union or European Commission. Neither the European Union nor the European Commission can be held responsible for them. The project on which this report is based was funded by the Federal Ministry of Education and Research under the funding code 16KIS2012. The responsibility for the content of this publication lies with the authors.



\section{Limitations}
Although we identify a \textcolor{ideal}{prescriptive} component influencing LLM outputs, the origin of these norms, whether they stem from the pre-training data, reinforcement learning from human feedback (RLHF), or some other aspect of model training remains under-explored. Further analysis is required to disentangle the contributions of training data versus fine-tuning techniques in shaping \textcolor{ideal}{prescriptive} tendencies in LLMs. Clarifying these origins could inform strategies to better control or mitigate unintended \textcolor{ideal}{prescriptive} biases in model outputs. The paper also does not explore the mechanism of how norms affect heuristics.

Furthermore, this work evaluates prototypicality in LLM similar to evaluation in human subjects. But, prototypicality in neural networks can be studied more closely using their representations. Though the prototypical analysis is stated as an initial exploration in the manuscript, it calls for further research in mechanistic analysis of how prototypes contain \textcolor{ideal}{prescriptive} norms and the possibility of steering and controlling of these norms in concept representations.

\section{Ethics and Risks}

This paper investigates the sampling heuristics of LLMs, revealing a \textcolor{ideal}{prescriptive} bias that may impact decision-making in real-world applications. While such biases could align outputs with certain normative expectations, they raise ethical concerns as there is no guarantee of such an alignment. This is particularly important in contexts like healthcare and policy-making, where fairness and transparency are critical. Understanding and mitigating these biases is essential to prevent unintended harm and ensure the responsible deployment of LLMs.

Furthermore, we hypothesise that this \textcolor{ideal}{prescriptive} norm acts as a foundational bias in other biases found in LLMs like gender, demography, etc which could be looked at through the lens of value. Since there are no guarantees on the \textcolor{ideal}{ideals} of LLMs, LLM sampled options can appear with different biases under different concepts/domains. This raises important ethical concerns, potentially leading to outputs that do not reflect (a) real-world norms or (b) diverse perspectives. Addressing influence of \textcolor{ideal}{prescriptive} norms is essential for developing transparent, reliable, and fair AI technologies, ensuring they contribute positively and ethically across various societal applications.

\bibliography{custom}

\begin{thebibliography}{38}
\providecommand{\natexlab}[1]{#1}

\bibitem[{Achiam et~al.(2023)Achiam, Adler, Agarwal, Ahmad, Akkaya, Aleman, Almeida, Altenschmidt, Altman, Anadkat et~al.}]{achiam2023gpt}
Josh Achiam, Steven Adler, Sandhini Agarwal, Lama Ahmad, Ilge Akkaya, Florencia~Leoni Aleman, Diogo Almeida, Janko Altenschmidt, Sam Altman, Shyamal Anadkat, et~al. 2023.
\newblock Gpt-4 technical report.
\newblock \emph{arXiv preprint arXiv:2303.08774}.

\bibitem[{Anthropic(2024)}]{anthropic2024claude}
AI~Anthropic. 2024.
\newblock The claude 3 model family: Opus, sonnet, haiku.
\newblock \emph{Claude-3 Model Card}.

\bibitem[{Barsalou(1985)}]{barsalou1985ideals}
Lawrence~W Barsalou. 1985.
\newblock Ideals, central tendency, and frequency of instantiation as determinants of graded structure in categories.
\newblock \emph{Journal of experimental psychology: learning, memory, and cognition}, 11(4):629.

\bibitem[{Bear et~al.(2020)Bear, Bensinger, Jara-Ettinger, Knobe, and Cushman}]{bear2020comes}
Adam Bear, Samantha Bensinger, Julian Jara-Ettinger, Joshua Knobe, and Fiery Cushman. 2020.
\newblock What comes to mind?
\newblock \emph{Cognition}, 194:104057.

\bibitem[{Bear and Knobe(2017)}]{bear2017normality}
Adam Bear and Joshua Knobe. 2017.
\newblock Normality: Part descriptive, part prescriptive.
\newblock \emph{Cognition}, 167:25--37.

\bibitem[{Bender et~al.(2021)Bender, Gebru, McMillan-Major, and Shmitchell}]{bender2021dangers}
Emily~M Bender, Timnit Gebru, Angelina McMillan-Major, and Shmargaret Shmitchell. 2021.
\newblock On the dangers of stochastic parrots: Can language models be too big?
\newblock In \emph{Proceedings of the 2021 ACM conference on fairness, accountability, and transparency}, pages 610--623.

\bibitem[{Brown et~al.(2020)Brown, Mann, Ryder, Subbiah, Kaplan, Dhariwal, Neelakantan, Shyam, Sastry, Askell et~al.}]{brown2020language}
Tom Brown, Benjamin Mann, Nick Ryder, Melanie Subbiah, Jared~D Kaplan, Prafulla Dhariwal, Arvind Neelakantan, Pranav Shyam, Girish Sastry, Amanda Askell, et~al. 2020.
\newblock Language models are few-shot learners.
\newblock \emph{Advances in neural information processing systems}, 33:1877--1901.

\bibitem[{Gallegos et~al.(2024)Gallegos, Rossi, Barrow, Tanjim, Kim, Dernoncourt, Yu, Zhang, and Ahmed}]{gallegos2023bias}
Isabel~O Gallegos, Ryan~A Rossi, Joe Barrow, Md~Mehrab Tanjim, Sungchul Kim, Franck Dernoncourt, Tong Yu, Ruiyi Zhang, and Nesreen~K Ahmed. 2024.
\newblock Bias and fairness in large language models: A survey.
\newblock \emph{Computational Linguistics}, 50(3):1097--1179.

\bibitem[{Gigerenzer and Gaissmaier(2011)}]{gigerenzer2011heuristic}
Gerd Gigerenzer and Wolfgang Gaissmaier. 2011.
\newblock Heuristic decision making.
\newblock \emph{Annual review of psychology}, 62(1):451--482.

\bibitem[{Gu et~al.(2025)Gu, Pang, Shen, and Cheng}]{gu2024llms}
Jia Gu, Liang Pang, Huawei Shen, and Xueqi Cheng. 2025.
\newblock \href {https://aclanthology.org/2025.coling-main.360/} {Do {LLM}s play dice? exploring probability distribution sampling in large language models for behavioral simulation}.
\newblock In \emph{Proceedings of the 31st International Conference on Computational Linguistics}, pages 5375--5390, Abu Dhabi, UAE. Association for Computational Linguistics.

\bibitem[{Guo et~al.(2025)Guo, Yang, Zhang, Song, Zhang, Xu, Zhu, Ma, Wang, Bi et~al.}]{guo2025deepseek}
Daya Guo, Dejian Yang, Haowei Zhang, Junxiao Song, Ruoyu Zhang, Runxin Xu, Qihao Zhu, Shirong Ma, Peiyi Wang, Xiao Bi, et~al. 2025.
\newblock Deepseek-r1: Incentivizing reasoning capability in llms via reinforcement learning.
\newblock \emph{arXiv preprint arXiv:2501.12948}.

\bibitem[{Hastings(2024)}]{hastings2024preventing}
Janna Hastings. 2024.
\newblock Preventing harm from non-conscious bias in medical generative ai.
\newblock \emph{The Lancet Digital Health}, 6(1):e2--e3.

\bibitem[{Hazra et~al.(2024)Hazra, Dos~Martires, and De~Raedt}]{hazra2023saycanpay}
Rishi Hazra, Pedro~Zuidberg Dos~Martires, and Luc De~Raedt. 2024.
\newblock Saycanpay: Heuristic planning with large language models using learnable domain knowledge.
\newblock In \emph{Proceedings of the AAAI Conference on Artificial Intelligence}, volume~38, pages 20123--20133.

\bibitem[{Jiang et~al.(2023)Jiang, Sablayrolles, Mensch, Bamford, Chaplot, Casas, Bressand, Lengyel, Lample, Saulnier et~al.}]{jiang2023mistral}
Albert~Q Jiang, Alexandre Sablayrolles, Arthur Mensch, Chris Bamford, Devendra~Singh Chaplot, Diego de~las Casas, Florian Bressand, Gianna Lengyel, Guillaume Lample, Lucile Saulnier, et~al. 2023.
\newblock Mistral 7b.
\newblock \emph{arXiv preprint arXiv:2310.06825}.

\bibitem[{Jiang et~al.(2024)Jiang, Sablayrolles, Roux, Mensch, Savary, Bamford, Chaplot, Casas, Hanna, Bressand et~al.}]{jiang2024mixtral}
Albert~Q Jiang, Alexandre Sablayrolles, Antoine Roux, Arthur Mensch, Blanche Savary, Chris Bamford, Devendra~Singh Chaplot, Diego de~las Casas, Emma~Bou Hanna, Florian Bressand, et~al. 2024.
\newblock Mixtral of experts.
\newblock \emph{arXiv preprint arXiv:2401.04088}.

\bibitem[{Jin and Rinard(2024)}]{jin2023evidence}
Charles Jin and Martin Rinard. 2024.
\newblock Emergent representations of program semantics in language models trained on programs.
\newblock In \emph{Forty-first International Conference on Machine Learning}.

\bibitem[{Kahneman(2011)}]{daniel2017thinking}
Daniel Kahneman. 2011.
\newblock Fast and slow thinking.
\newblock \emph{Allen Lane and Penguin Books, New York}.

\bibitem[{Lampinen et~al.(2024)Lampinen, Dasgupta, Chan, Sheahan, Creswell, Kumaran, McClelland, and Hill}]{dasgupta2022language}
Andrew~K Lampinen, Ishita Dasgupta, Stephanie~CY Chan, Hannah~R Sheahan, Antonia Creswell, Dharshan Kumaran, James~L McClelland, and Felix Hill. 2024.
\newblock Language models, like humans, show content effects on reasoning tasks.
\newblock \emph{PNAS nexus}, 3(7):pgae233.

\bibitem[{Li et~al.(2023)Li, Hopkins, Bau, Vi{\'e}gas, Pfister, and Wattenberg}]{li2022emergent}
Kenneth Li, Aspen~K Hopkins, David Bau, Fernanda Vi{\'e}gas, Hanspeter Pfister, and Martin Wattenberg. 2023.
\newblock Emergent world representations: Exploring a sequence model trained on a synthetic task.
\newblock \emph{ICLR}.

\bibitem[{Li et~al.(2025)Li, Zhang, Zhang, Zhang, Liu, Yao, Xu, Zheng, Wang, Chen et~al.}]{li2025system}
Zhong-Zhi Li, Duzhen Zhang, Ming-Liang Zhang, Jiaxin Zhang, Zengyan Liu, Yuxuan Yao, Haotian Xu, Junhao Zheng, Pei-Jie Wang, Xiuyi Chen, et~al. 2025.
\newblock From system 1 to system 2: A survey of reasoning large language models.
\newblock \emph{arXiv preprint arXiv:2502.17419}.

\bibitem[{Mattar and Daw(2018)}]{mattar2018prioritized}
Marcelo~G Mattar and Nathaniel~D Daw. 2018.
\newblock Prioritized memory access explains planning and hippocampal replay.
\newblock \emph{Nature neuroscience}, 21(11):1609--1617.

\bibitem[{Mattar and Lengyel(2022)}]{mattar2022planning}
Marcelo~G Mattar and M{\'a}t{\'e} Lengyel. 2022.
\newblock Planning in the brain.
\newblock \emph{Neuron}, 110(6):914--934.

\bibitem[{McKenzie et~al.(2023)McKenzie, Lyzhov, Pieler, Parrish, Mueller, Prabhu, McLean, Kirtland, Ross, Liu et~al.}]{mckenzie2023inverse}
Ian~R McKenzie, Alexander Lyzhov, Michael Pieler, Alicia Parrish, Aaron Mueller, Ameya Prabhu, Euan McLean, Aaron Kirtland, Alexis Ross, Alisa Liu, et~al. 2023.
\newblock Inverse scaling: When bigger isn't better.
\newblock \emph{arXiv preprint arXiv:2306.09479}.

\bibitem[{Murphy(2004)}]{murphy2004big}
Gregory Murphy. 2004.
\newblock \emph{The big book of concepts}.
\newblock MIT press.

\bibitem[{Omiye et~al.(2023)Omiye, Lester, Spichak, Rotemberg, and Daneshjou}]{omiye2023large}
Jesutofunmi~A Omiye, Jenna~C Lester, Simon Spichak, Veronica Rotemberg, and Roxana Daneshjou. 2023.
\newblock Large language models propagate race-based medicine.
\newblock \emph{NPJ Digital Medicine}, 6(1):195.

\bibitem[{OpenAI(2024)}]{openai2024reasoning}
OpenAI. 2024.
\newblock \href {https://openai.com/index/learning-to-reason-with-llms/} {Learning to reason with llms}.
\newblock Accessed: 2025-05-30.

\bibitem[{Phillips et~al.(2019)Phillips, Morris, and Cushman}]{phillips2019we}
Jonathan Phillips, Adam Morris, and Fiery Cushman. 2019.
\newblock How we know what not to think.
\newblock \emph{Trends in cognitive sciences}, 23(12):1026--1040.

\bibitem[{Ross et~al.(2023)Ross, Gl{\u{a}}veanu, and Baumeister}]{ross2023new}
Wendy Ross, Vlad Gl{\u{a}}veanu, and Roy~F Baumeister. 2023.
\newblock The new science of possibility.
\newblock \emph{Possibility Studies Society}, 1(4):399--403.

\bibitem[{Shah et~al.(2023)Shah, Equi, Osi{\'n}ski, Xia, Ichter, and Levine}]{shah2023navigation}
Dhruv Shah, Michael~Robert Equi, B{\l}a{\.z}ej Osi{\'n}ski, Fei Xia, Brian Ichter, and Sergey Levine. 2023.
\newblock Navigation with large language models: Semantic guesswork as a heuristic for planning.
\newblock In \emph{Conference on Robot Learning}, pages 2683--2699. PMLR.

\bibitem[{Silver et~al.(2016)Silver, Huang, Maddison, Guez, Sifre, Van Den~Driessche, Schrittwieser, Antonoglou, Panneershelvam, Lanctot et~al.}]{silver2016mastering}
David Silver, Aja Huang, Chris~J Maddison, Arthur Guez, Laurent Sifre, George Van Den~Driessche, Julian Schrittwieser, Ioannis Antonoglou, Veda Panneershelvam, Marc Lanctot, et~al. 2016.
\newblock Mastering the game of go with deep neural networks and tree search.
\newblock \emph{nature}, 529(7587):484--489.

\bibitem[{Smith and Medin(1981)}]{smith1981categories}
Edward~E Smith and Douglas~L Medin. 1981.
\newblock \emph{Categories and concepts}.
\newblock Harvard University Press.

\bibitem[{Suri et~al.(2024)Suri, Slater, Ziaee, and Nguyen}]{suri2023large}
Gaurav Suri, Lily~R Slater, Ali Ziaee, and Morgan Nguyen. 2024.
\newblock Do large language models show decision heuristics similar to humans? a case study using gpt-3.5.
\newblock \emph{Journal of Experimental Psychology: General}, 153(4):1066.

\bibitem[{Thirunavukarasu et~al.(2023)Thirunavukarasu, Ting, Elangovan, Gutierrez, Tan, and Ting}]{thirunavukarasu2023large}
Arun~James Thirunavukarasu, Darren Shu~Jeng Ting, Kabilan Elangovan, Laura Gutierrez, Ting~Fang Tan, and Daniel Shu~Wei Ting. 2023.
\newblock Large language models in medicine.
\newblock \emph{Nature medicine}, 29(8):1930--1940.

\bibitem[{Touvron et~al.(2023)Touvron, Lavril, Izacard, Martinet, Lachaux, Lacroix, Rozi{\`e}re, Goyal, Hambro, Azhar et~al.}]{touvron2023llama}
Hugo Touvron, Thibaut Lavril, Gautier Izacard, Xavier Martinet, Marie-Anne Lachaux, Timoth{\'e}e Lacroix, Baptiste Rozi{\`e}re, Naman Goyal, Eric Hambro, Faisal Azhar, et~al. 2023.
\newblock Llama: Open and efficient foundation language models.
\newblock \emph{arXiv preprint arXiv:2302.13971}.

\bibitem[{Wan et~al.(2023)Wan, Pu, Sun, Garimella, Chang, and Peng}]{wan2023kelly}
Yixin Wan, George Pu, Jiao Sun, Aparna Garimella, Kai-Wei Chang, and Nanyun Peng. 2023.
\newblock “kelly is a warm person, joseph is a role model”: Gender biases in llm-generated reference letters.
\newblock In \emph{Findings of the Association for Computational Linguistics: EMNLP 2023}, pages 3730--3748.

\bibitem[{Wei et~al.(2022)Wei, Wang, Schuurmans, Bosma, Xia, Chi, Le, Zhou et~al.}]{wei2022chain}
Jason Wei, Xuezhi Wang, Dale Schuurmans, Maarten Bosma, Fei Xia, Ed~Chi, Quoc~V Le, Denny Zhou, et~al. 2022.
\newblock Chain-of-thought prompting elicits reasoning in large language models.
\newblock \emph{Advances in Neural Information Processing Systems}, 35:24824--24837.

\bibitem[{Yao et~al.(2023)Yao, Yu, Zhao, Shafran, Griffiths, Cao, and Narasimhan}]{yao2023tree}
Shunyu Yao, Dian Yu, Jeffrey Zhao, Izhak Shafran, Tom Griffiths, Yuan Cao, and Karthik Narasimhan. 2023.
\newblock Tree of thoughts: Deliberate problem solving with large language models.
\newblock \emph{Advances in neural information processing systems}, 36:11809--11822.

\bibitem[{Zack et~al.(2024)Zack, Lehman, Suzgun, Rodriguez, Celi, Gichoya, Jurafsky, Szolovits, Bates, Abdulnour et~al.}]{zack2024assessing}
Travis Zack, Eric Lehman, Mirac Suzgun, Jorge~A Rodriguez, Leo~Anthony Celi, Judy Gichoya, Dan Jurafsky, Peter Szolovits, David~W Bates, Raja-Elie~E Abdulnour, et~al. 2024.
\newblock Assessing the potential of gpt-4 to perpetuate racial and gender biases in health care: a model evaluation study.
\newblock \emph{The Lancet Digital Health}, 6(1):e12--e22.

\end{thebibliography}

\clearpage

\section*{Appendix}
\label{sec:appendix}

\appendix
\setcounter{section}{0}

\section{Glossary} \label{sec:glossary}
\begin{itemize}
    \item Sampling: The process of selecting one or more outcomes from a set of possible options based on some probability distribution. In the context of the manuscript sampling is defined the process by which the LLM probabilistically selects outputs from a distribution of potential options. 
    \item Heuristics: Heuristics generally refer to mental shortcuts or rule-based approximations. In the context of the manuscript, it refer to empirically derived rules the LLM employ to streamline sample generation processes by approximating deliberate outcomes without incurring the cost of exhaustive search through decision branches.
    \item Prescriptive component: The prescriptive component of a concept reflects an implicit ideal or normative standard of the concept encoded within the cognitive agent or the model. In human cognition, it reflects the value of the concept and can manifest as moral, cultural, or goal-oriented biases in decision-making. In LLMs, the prescriptive component seems to emerge from patterns in training data and RLHF, influencing outputs to align with an implicit notion of "ideal" rather than just statistical norms. The notion of an "ideal" in the LLM need not align with human values. 
    \item Descriptive component: The descriptive component refers to observed patterns that define what is typical or statistically frequent in a given concept. In LLMs, it corresponds to the underlying statistical probability distribution learned from pretraining data for each concept, reflecting common word sequences and structures.
    
    \item Prototpye: A prototype is the most representative example of a concept. In humans, prototype is the cognitive “average” of a category—a mental representation that encapsulates the most typical features shared by its members. It serves as a benchmark against which new instances are compared to decide if they belong to that category. Prototypes has been shown to useful in ML for understanding how well a concept generalize across scenarios.
    
    \item System-1: System-1 refers to a mode of decision-making characterized by fast, automatic, and intuitive processing that relies on heuristics rather than explicit reasoning. This enables rapid decision-making often at the cost of accuracy and depth. In human cognition, System-1 is responsible for routine tasks, immediate responses, and heuristic-driven judgments, often without conscious deliberation. 
    
    In LLMs, System-1-like behavior corresponds to the probabilistic selection of tokens based on learned statistical patterns, without explicit multi-step reasoning or deliberation. This results in fluent but potentially biased or heuristic-driven outputs, similar to human cognitive shortcuts.
    
    \item System-2: System-2 is a slow, deliberate, and analytical mode of thinking that requires cognitive effort and logical reasoning. In human cognition, it is responsible for problem-solving and long-term planning. In LLMs, System-2-like behavior is induced through structured prompting techniques, such as chain-of-thought reasoning, where intermediate steps are explicitly modeled.
    
    \item Value-system: A value system is a structured hierarchical framework of beliefs, morals/principles, and standards that guide how individuals or groups determine what is important, good, or desirable. It influences decisions, behavior, and priorities by providing a set of criteria against which actions and outcomes are judged. In LLMs, a value-system is not explicitly encoded but emerges through training data biases, reinforcement learning objectives, and alignment mechanisms that shape the model’s preferences for certain types of sampling outputs over others.
    
    \item Normality: Normality of the concept in simple words is what is considered normal of that concept. It is defined by the set of observed behaviors or patterns of elements of a concept that align with established or typical standards of the concept. Normality in humans is found to be a cognitive representation that integrates descriptive norms (statistical regularities—what is common or average) and prescriptive norms (idealized expectations—what is good, desirable, or appropriate). We find that LLMs concept of normality and what is normal also incorporates both these dimensions indicating that prototypical representations are biased by value potentially raising ethical issues in downstream tasks.

    \item Concept: For LLM a concept refers to an abstract representation formed through statistical associations in training data, capturing relationships between words, phrases, and ideas in high-dimensional latent space. Unlike human-defined categories, LLM concepts emerge from probabilistic patterns of usage rather than explicit rule-based definitions, allowing generalization across contexts.
    
    \item  Exemplar : An exemplar is defined as a specific instance or example of a concept that people use to represent that concept in their minds. Unlike prototypes, which can be abstracted averages of category members, exemplars are concrete instances stored in memory.
    In the context of the paper, an exemplar serves as a specific, descriptive representation of an example of a concept that an LLM evaluates based on statistical norms (descriptive components) and idealized values (prescriptive components). In this work we find how LLMs, like humans, assess exemplars by considering not just their statistical frequency within a category but also the implicit values associated with them.

\end{itemize}

\section{Compute Resources and Licenses } \label{sec:compute_resources}

We use API to access the LLMs. We do not load the models locally. For GPT we use the Open-AI API. The API used for open source models shall be revealed once the double-blind is no longer valid. When utilizing large language models such as GPT (OpenAI), Claude (Anthropic), LLaMA (Meta), and Mistral in scientific research, we cite the respective models. Each model's terms dictate its permissible uses, including conditions for research, publication, and potential downstream applications. To ensure compliance, we have reviewed and adhered to these licenses in the preparation of this work.

LLaMA (Meta) is provided under a research license, allowing its application in academic work. Its deployment in this study aligns with these conditions, with clear citing of model. Similarly, Mistral models, released under permissive licenses, offer significant flexibility for research. Attribution requirements outlined in these licenses have been met, ensuring compliance with their terms. More details on services that host open sourced models will be revealed after the effect of double blind policy stops applying. In summary, this work complies with all licensing and usage policies of the cited models. Attribution is provided as required, and the use of these tools is disclosed to maintain transparency and reproducibility in line with the standards of the research community.

\section{Understanding biases of LLMs}
\label{realted_work_app}

Previous work on LLM predominantly evaluates biases with respect to social concepts like gender, race and popularity. There has also been investigation of biases in aspects like language style and lexical content \cite{wan2023kelly}. Gallegos \etal gives a comprehensive survey of these works and presents a taxonomy of biases \cite{gallegos2023bias}. This taxonomy aligns with how humans attribute meanings to these biases and their impact on society. The biases of LLM have also been studied in the context of specific fields and applications like health care ~\cite{omiye2023large,zack2024assessing,thirunavukarasu2023large,hastings2024preventing}. These studies do not go beyond the human taxonomy of biases to explore fundamental biases that, in turn, manifest in real-world applications.

Biases in System-1 outputs significantly influence System-2 processes because the latter often depend on the former as a prior in decision-making. For instance, in AlphaGo~\citep{silver2016mastering}, the Monte Carlo Tree Search (MCTS) algorithm (a System-2 process) relies on estimates from a neural network (System-1) to limit the search space. Similarly, in frameworks like Tree of Thoughts (ToT)~\citep{yao2023tree}, LLMs generate initial samples that a symbolic solver refines, assuming that the LLM provides a useful prior for the problem solver. Understanding and explaining system-1 biases are pivotal to making system-2 based real world systems.

\begin{figure*}[htbp]
    \centering
    \includegraphics[width=0.8\linewidth]{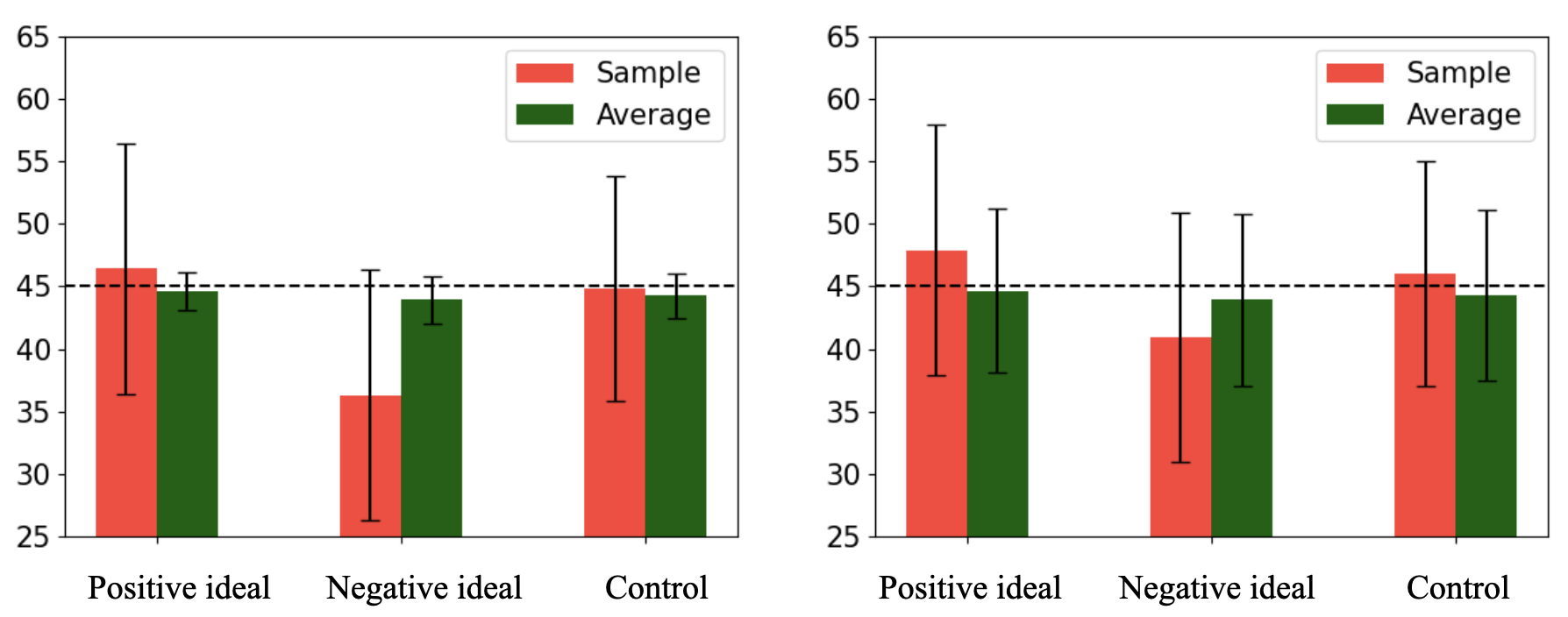}
    \caption{Estimates of the \textcolor{average}{average} amount of glubbing (green) and mean of \textcolor{sample}{samples} (red)  for the unimodal (left) and bimodal (right) conditions from the experiment~\ref{exp:glubbing}. The true average (mean of input distribution) is presented is also shown in dashed black lines.}
    \label{fig:glubbing}
    \vspace{-0.5cm}
\end{figure*}

\begin{figure*}[htbp]
    \centering
    \begin{subfigure}[t]{0.38\textwidth}
        \centering
        \includegraphics[width=\linewidth]{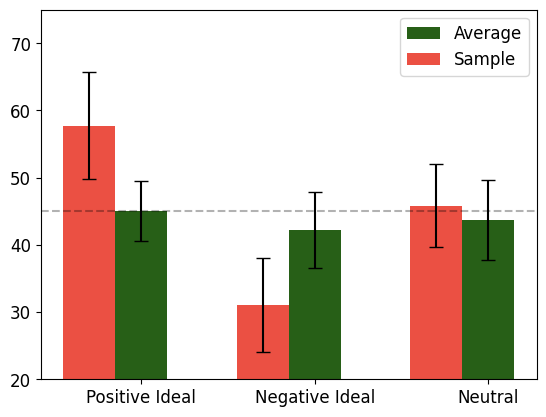}
    \end{subfigure}
    \hspace{0.02\textwidth} 
    \begin{subfigure}[t]{0.38\textwidth}
        \centering
        \includegraphics[width=\linewidth]{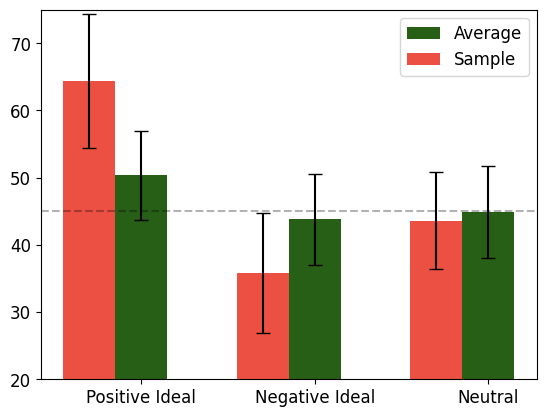}
    \end{subfigure}
    \caption{Estimates of the \textcolor{average}{average} amount of flubbing (green) and the mean of \textcolor{sample}{samples} (red) for the unimodal (left) and bimodal (right) conditions from the human experiment ~\citep{bear2020comes}. The true average (mean of the input distribution) is shown with dashed black lines.}
    \label{fig:combined_glubbing}
    \vspace{-0.5cm}
\end{figure*}

\section{Sampling on novel concept: human experiment}
\label{sec:flubbing}
A total of 1,200 participants were assigned across six conditions in a $2 \times 3$ pre-registered design. The experiment manipulated the statistical distribution of new concept flubbing amounts (unimodal vs. bimodal) and prescriptive value (high, low, or neutral ideal). Specifically, the flubbing amounts were drawn from:

\begin{itemize}
    \item \textbf{Unimodal distribution:} $\mu = 45$, $\sigma = 15$
    \item \textbf{Bimodal distribution:} $\mu_1 = 35$, $\mu_2 = 75$, $\sigma = 5$
\end{itemize}

For the prescriptive value conditions:

\begin{itemize}
    \item \textbf{High ideal:} Flubbing amounts greater than 80 minutes were ideal (A+), while amounts less than 20 minutes received the lowest grade (D-).
    \item \textbf{Low ideal:} Amounts less than 20 minutes were ideal (A+), and those above 80 were discouraged (D-).
    \item \textbf{Intermediate ideal:} The ideal amount of flubbing was set to 50 minutes, and grades were linearly scaled based on deviation from 50.
\end{itemize}

After viewing 100 amounts of flubbing paired with health grades, participants were asked to report the first number of minutes of flubbing that came to mind. The results showed:

\begin{itemize}
    \item Participants' sample judgments significantly differed from their estimates of the average flubbing amount. For the \textbf{low ideal} condition, the paired t-test yielded $t(331) = 11.98, p < .001$. For the \textbf{high ideal} condition, the paired t-test was $t(293) = 16.55, p < .001$. 
    \item In the \textbf{intermediate ideal} condition, sample judgments and estimates of average flubbing did not significantly diverge, $t(318) = 0.085, p = .93$.
\end{itemize}

In analyzing the computational models, the \textit{softmax model} provided the best fit across conditions when compared to other models, such as the additive and multiplicative models. The \textit{softmax model} predicted participants’ sample judgments as a combination of statistical probability \textit{$C_{a}$} (distribution average) and prescriptive value \textit{$C_v$}. The product of these factors explained the distribution of flubbing amounts that came to mind. 

\[
P(x) = \frac{e^{C_v(x)}}{\sum e^{C_v(x')}} \times C_{\mu}(x)
\]

The mean sample judgments is significantly influenced by the prescriptive values \textit{$C_v$}, with deviations from the true average \textit{$C_{\mu}$}. The differences between sample judgments and participants' estimates of average flubbing were highly significant in both the low ideal condition ($ p < .001$) and the high ideal condition ($ p < .001$). No significant difference was found in the intermediate ideal condition ($ p = .93$). These results suggest that participants were strongly influenced by prescriptive values in their judgments. The results are shown in Figure \ref{fig:combined_glubbing}.

This experiment was replicated in this paper and we return similar results where the LLM also shows strong influence of prescriptive values as shown in Figure \ref{fig:glubbing}. The  similarity of the two figures strongly validate the proposed theory-the sampling heuristics of LLM and humans allign.

\section{Sampling in relation to existing concepts in humans}
\label{sec:exp1_humans}

In this section, we present the experiment~\ref{implicit} on the same concepts and using the same prompt as in prior work in humans by~\citet{bear2020comes}. The results for LLM are shown in Table \ref{tab:implicit value bias} and the results for humans in the same concepts are shown in Table \ref{tab:bear_paper}. Comparing this result with the human studies, as shown in Appendix~\ref{sec:exp1_humans}, we observe that the LLM often gives a `strictly ideal' value when queried for \textit{$I(C)$}. That is, when a similar question is asked to human test subjects, the number of concepts for which the \textcolor{ideal}{ideal} value is zero is only one. On the other hand, the LLM gives zero for \textit{$I(C)$} for 19 concepts (nearly half the time). For instance, the human gives the \textcolor{ideal}{ideal} percentage of `high school students underage drinking' as 13.71\%, while the LLM gives \textit{$I(C)$} as zero for this concept, showing LLMs, for a lot of concepts, have a notion of stricter ideality compared to the more noisy ideal ratings we seem to observe across humans.

We also repeat this experiment for temperature zero as shown in Table \ref{tab:exp1_implicit_value_temp_zero} in Section \ref{sec:exp1_temp_zero}, and observe similar results. We get the following results with other LLMs with default temperatures: Llama-3-7b (binomial $p$ = 0.003), Mixtral-8x7B (binomial $p$ = 0.05), GPT3.5-turbo (binomial $p$ < 0.001),  Claude (binomial $p$ < 0.001), Mistral (binomial $p$ = 0.0019).

\begin{figure}[h]
    \centering
    \includegraphics[width=\linewidth]{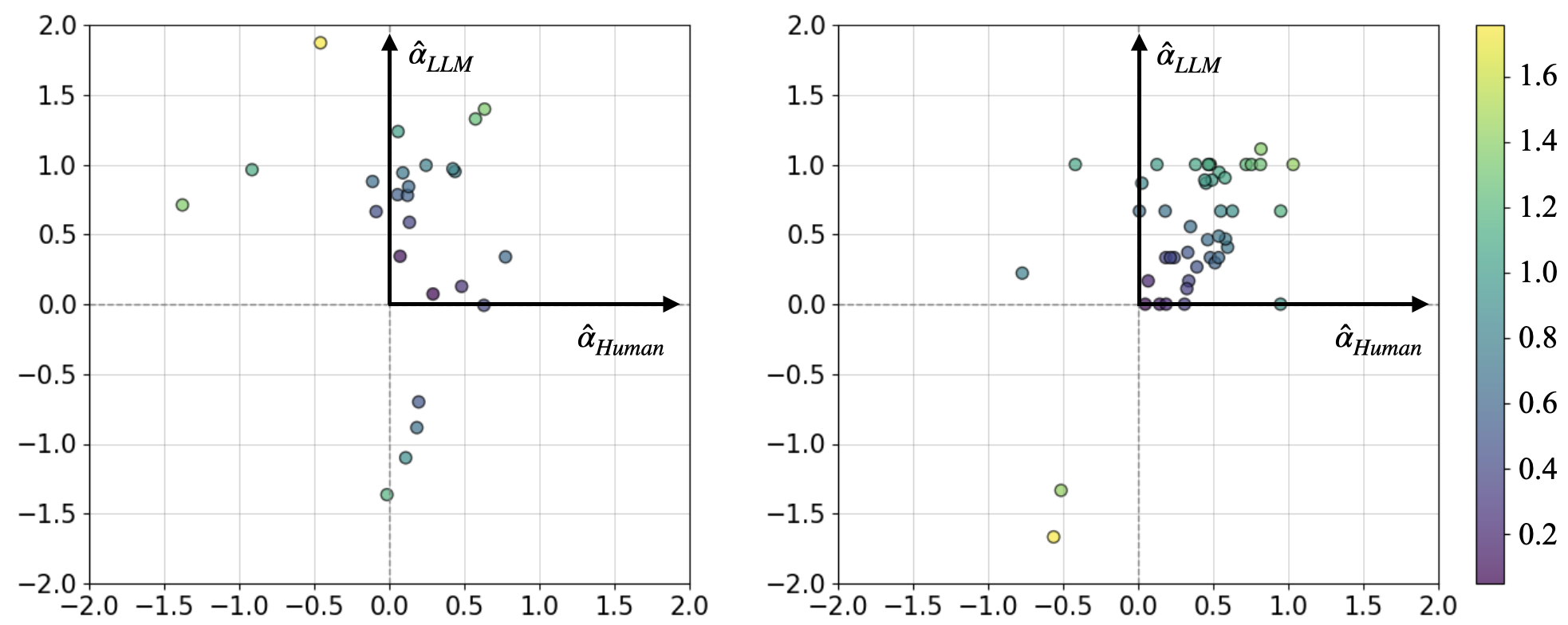}
    \caption{Comparing human and LLM on the prototype experiment and sampling on existing concepts. Figure on the left compares from results in Experiment 2 showing some misalignment between LLM and human results due to differences arising in the prescriptive component. Figure on the right compares LLM and human results from Experiment 3 showing more correlation in prototypical concept ratings.}
    \label{fig:alpha_plot_full}
\end{figure}

We present a scatter plot of \textit{$\hat{\alpha}$} values for LLMs and humans in Figure~\ref{fig:alpha_plot_full}. We see that although the LLM has a strong  \textcolor{ideal}{prescriptive} component based on its implicit value associated with each concept, its amplitude of shift towards \textcolor{ideal}{prescriptive} norm does not correlate with that of humans (Pearson correlation of -0.02). \textbf{This makes the study of prescriptive norms in LLMs more significant as they might not manifest in samples like it does in humans}. Comparing \textit{$\hat{\alpha}$} of humans and the LLM for experiment~\ref{prototype} shows a higher alignment in the shift (Figure~\ref{fig:alpha_plot_full}). Here the Pearson correlation of $\hat{\alpha}_\textit{human}$ and $\hat{\alpha}_\textit{LLM}$ is 0.33.

\begin{table*}[h]
\tiny
\centering
\begin{tabular}{>{\raggedright\arraybackslash}p{4cm}ccc>{\raggedright\arraybackslash}p{4cm}ccc}
\toprule
\textbf{Domain} & \textcolor{average}{\textbf{Average}} & \textcolor{ideal}{\textbf{Ideal}} & \textcolor{sample}{\textbf{Sample}} & \textbf{Domain} & \textcolor{average}{\textbf{Average}} & \textcolor{ideal}{\textbf{Ideal}} & \textcolor{sample}{\textbf{Sample}} \\
\midrule
Hours TV/day & 3.38 & 1.63 & 2.87 & Drinks frat bro consume/wknd & 11.12 & 6.63 & 15.64 \\
Sugary drinks/wk & 9.17 & 2.41 & 5.91 & Times honk at drivers/wk & 2.67 & 0.72 & 2.53 \\ 
Hours Exercise/wk & 4.00 & 5.58 & 6.33 & Mins on social media/day & 60.57 & 35.40 & 59.10 \\
Cals consumed/day & 2225.91 & 1900.00 & 1859.24 & Times parent punishes child/month & 6.58 & 2.28 & 3.25 \\
Servings Fruits \& veggies/month & 40.00 & 94.96 & 39.16 & Miles walked/wk & 9.79 & 12.96 & 9.96 \\
Lies told/wk & 9.57 & 1.17 & 8.44 & \% people drive drunk & 11.30 & 1.23 & 9.45 \\
Mins late for appointment & 14.22 & 3.04 & 13.6 & Times cheat on partner in life & 1.52 & 0.00 & 1.73 \\
Books read/yr & 7.22 & 17.40 & 8.45 & Times snooze alarm/day & 2.13 & 0.76 & 1.98 \\
Romantic partners in life & 6.09 & 5.77 & 8.06 & Parking tickets/yr & 1.67 & 0.04 & 1.37 \\
Country's international conflicts/decade & 11.67 & 1.36 & 4.15 & Times car wash/yr & 10.77 & 12.85 & 11.31 \\
Dollars cheated on taxes & 437.45 & 82.0 & 350.32 & Cups coffee/day & 2.21 & 1.84 & 2.72 \\
\% students cheat on HS exam & 33.00 & 2.17 & 19.50 & Desserts/wk & 3.85 & 2.92 & 4.04 \\
Times checking phone/day & 28.57 & 7.68 & 16.57 & Loads of laundry/wk & 3.42 & 2.70 & 3.75 \\
Mins waiting on phone for customer service & 20.21 & 3.88 & 13.29 & \% HS students underage drink & 35.81 & 13.71 & 32.96 \\
Times called parents/month & 5.00 & 5.50 & 7.04 & \% students lying website & 50.56 & 13.40 & 47.20 \\
Times clean home/month & 5.78 & 4.35 & 6.24 & Servings carbs/day & 62.43 & 16.13 & 33.23 \\
Times computer crashes/wk & 3.07 & 0.12 & 1.14 & Txt msgs sent/day & 27.18 & 12.88 & 18.10 \\
\% HS dropouts & 10.67 & 1.29 & 11.49 & Times lose temper/wk & 2.60 & 0.56 & 2.20 \\
\% middle schoolers bullied & 17.59 & 0.81 & 19.46 & Times swearing/day & 8.69 & 5.88 & 11.26 \\
Hrs sleep/night & 6.69 & 7.84 & 7.32 & & & & \\
\bottomrule
\end{tabular}
\caption{Comparison of Average, Ideal, and Sample Data in various Domains \citep{bear2020comes}. The table shows human response sampling having a \textcolor{ideal}{prescriptive} norm component across concepts.}
\label{tab:bear_paper}
\end{table*}

\begin{table*}[t]

\tiny

\begin{tabular}{>{\raggedright\arraybackslash}p{3.9cm}rrr>{\raggedright\arraybackslash}p{3.1cm}rrr}
\toprule
\textbf{concept} & \textbf{Average} & \textbf{Ideal} & \textbf{Sample} & \textbf{concept} & \textbf{Average} & \textbf{Ideal} & \textbf{Sample} \\
\midrule
\textbf{Hours of TV in a day} & \textbf{3.36} & \textbf{1.85} & \textbf{3.25} & \textbf{Drinks in a frat weekend} & \textbf{12.87} & \textbf{7.87} & \textbf{2.65} \\
\textbf{Sugary drinks in a week} & \textbf{6.53} & \textbf{0.00} & \textbf{5.70} & \% people in a city driving drunk & 1.38 & 0.00 & 2.60 \\
Hours exercising in a week & 7.45 & 8.40 & 4.55 & Times to cheat on a partner in life & 1.28 & 0.00 & 15.29 \\
\textbf{Lies in a week} & \textbf{8.46} & \textbf{0.00} & \textbf{3.50} & Times to hit snooze on an alarm/day & 1.60 & 0.10 & 3.25 \\
\textbf{Calories in a day} & \textbf{2400.00} & \textbf{2000.00} & \textbf{3.70} & Parking tickets in a year & 2.05 & 0.00 & 5.50 \\
Servings of fruits and vegetables in a month & 69.93 & 108.00 & 18.00 & \textbf{Times to get car washed in a year} & \textbf{12.02} & \textbf{12.00} & \textbf{3.34} \\
\textbf{Number of minutes late for an appointment} & \textbf{14.36} & \textbf{0.00} & \textbf{3.10} & \textbf{Cups of coffee to drink in a day} & \textbf{1.85} & \textbf{2.80} & \textbf{2.52} \\
\textbf{Romantic partners in a lifetime} & \textbf{7.20} & \textbf{3.87} & \textbf{3.55} & \textbf{Loads of laundry to do in a week} & \textbf{2.06} & \textbf{3.15} & \textbf{4.10} \\
International conflicts in a decade & 1.07 & 0.00 & 3.55 & \textbf{\% of adults in a city smoking} & \textbf{20.38} & \textbf{0.00} & \textbf{4.50} \\
\textbf{Dollars to cheat on taxes} & \textbf{508.00} & \textbf{0.00} & \textbf{2.88} & \textbf{ \% of students drinking underage} & \textbf{32.55} & \textbf{0.00} & \textbf{5.15} \\
\textbf{ \% of students cheating on an exam} & \textbf{67.30} & \textbf{0.00} & \textbf{3.35} & \textbf{ \% of people lying on a dating site} & \textbf{55.06} & \textbf{0.00} & \textbf{3.27} \\
\textbf{Times to check a phone in a day} & \textbf{79.35} & \textbf{22.24} & \textbf{3.60} & Servings of carbohydrates in a day & 4.57 & 139.50 & 3.45 \\
\textbf{Min waiting on phone for customer service} & \textbf{11.30} & \textbf{3.10} & \textbf{3.35} & \textbf{Text messages to send in a day} & \textbf{94.00} & \textbf{34.50} & \textbf{10.90} \\
Times for a computer to crash in a week & 0.55 & 0.00 & 3.80 & Times to lose temper in a week & 3.50 & 0.00 & 5.95 \\
\textbf{\% of students dropping out of school} & \textbf{8.31} & \textbf{0.00} & \textbf{2.80} & \textbf{Times to swear in a day} & \textbf{80.00} & \textbf{0.00} & \textbf{2.97} \\
\textbf{\% of students being bullied in middle school} & \textbf{27.57} & \textbf{0.00} & \textbf{3.35} & \textbf{Times honk at drivers in a week} & \textbf{3.73} & \textbf{0.00} & \textbf{2.45} \\
Hours of sleep in a night & 7.40 & 7.70 & 3.20 & \textbf{Mins on social media in a day} & \textbf{144.10} & \textbf{30.00} & \textbf{3.05} \\
\textbf{Times parent punishes child in a month} & \textbf{4.99} & \textbf{0.00} & \textbf{3.30} & Miles walked in a week & 21.00 & 20.65 & 44.50 \\
\bottomrule
\end{tabular}
\caption{Comparison of \textcolor{average}{average}, \textcolor{ideal}{ideal}, and \textcolor{sample}{sample} data in various concepts, the concepts exhibiting \textcolor{ideal}{prescriptive norm} is in bold which makes up a significant number.}
\label{tab:implicit value bias}
\end{table*}

\section{Motivation for evaluating prototypes}
\label{mot_prts}

Barselou et al \citep{barsalou1985ideals} state that ideals may determine a concept's graded structure in one context, while central tendency may determine a different graded structure in another. In other words, when sampling, humans wouldn't use both \textcolor{ideal}{prescriptive} and \textcolor{average}{descriptive} prototypical ratings in the same context. But, Bear et al \citep{bear2017normality} show that human concepts have both components in the same context in a unified representation, providing an insight into how humans think about concepts, and our notion of normality is in fact both \textcolor{ideal}{prescriptive} and \textcolor{average}{descriptive}. When we try to rate a normal teacher, we include both \textcolor{ideal}{prescriptive} and \textcolor{average}{descriptive} components in the same context.

Given the two different theories, we test this in LLMs. Previous experiments in this paper show that LLMs, when sampling from innumerable options, use both \textcolor{ideal}{prescriptive} and \textcolor{average}{descriptive} norms as a heuristic in the same context akin to a unified representation. We show similar results of how prototypicality rating also has the same unified representation of both \textcolor{ideal}{prescriptive} and \textcolor{average}{descriptive} norms in the same context. We consider this experiment as an initial foray into how representations of prototypes drive cognitive biases. More work needs to be done to understand where these representative prototypes which have prescriptive norms exhibit unfavorably biased decision making.

Consider category 4 Exemplar 6 of Grandmother ``A 55-year-old woman who likes to party a lot and go out with her friends to casinos and rock concerts. Enjoys playing sports with her grandchildren" (Appendix ~\ref{sec:exp3_results}). This example of a grandmother has a lower \textcolor{ideal}{ideal} rating of 5.50 compared to other examples of the category. This is also reflected in the relative lower value of composite example rating (4.5), illustrating that non traditional prototypes are seen less ideally. Similar examples can been be seen in the table in Appendix \ref{sec:exp3_prompts}.

This implicit bias and punishing of non traditional prototypes has severe implications on tasks where LLM is asked to pick candidates whether it be for academic admissions or hiring processes. Another aspect this bias plays out is between the Exemplar 1 and Exemplar 2 of the Grandmother category. Even though Exemplar 2 has lesser \textcolor{average}{average} rating compared to Exemplar 1, having a more \textcolor{ideal}{ideal} rating makes it a better example of a grandmother compared to Exemplar 2 illustrating LLMs notion of concepts has a prescriptive norm component.

\begin{figure*}[t]
    \centering
    \includegraphics[width=\linewidth]{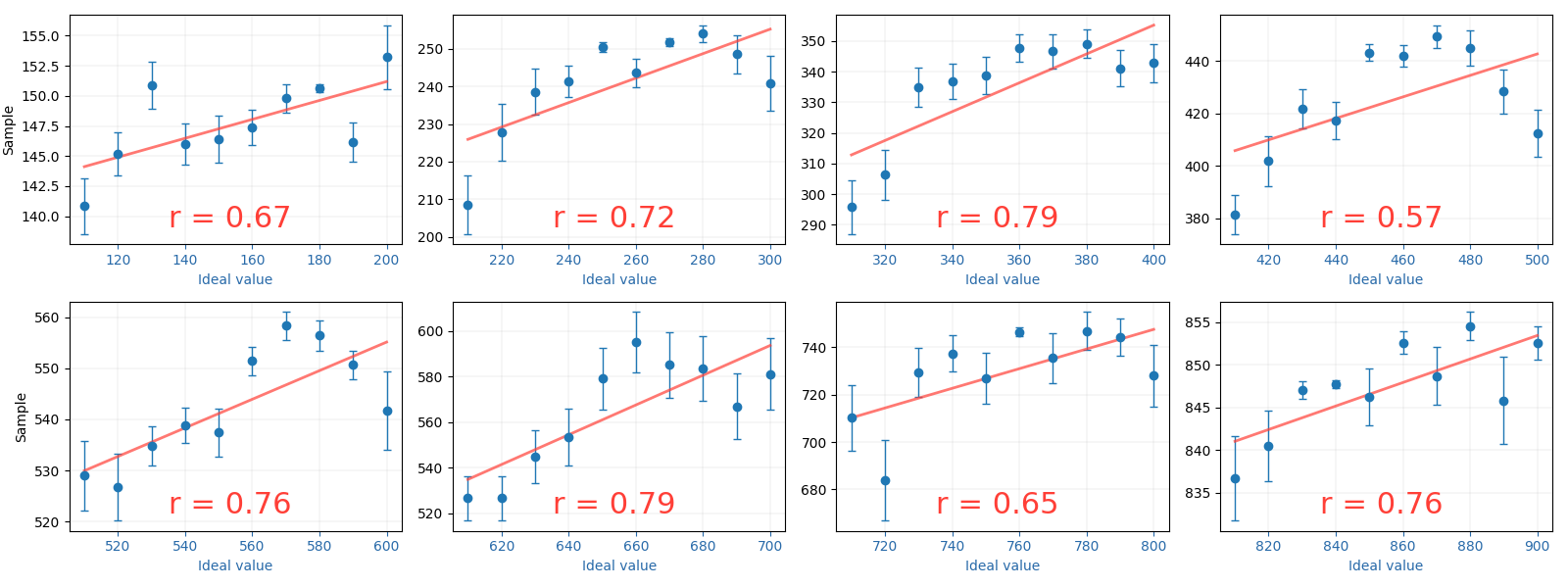}
    \caption{The figure shows the influence of the two components, showing strong evidence for the proposed theory. For each of the \textit{$C_{\mu}$}, changing \textit{$C_v$} clearly gives a shift in \textit{$S(C)$}. This show prescriptive norm has a stong influence on sampling across statistics. The vice versa is also true, given the \textit{$S(C)$} clearly changes with change of \textit{$C_\mu$}. The slope value across plots shows that effect of \textcolor{ideal}{prescriptive} norm is remarkably consistent.}
    \label{fig:beuty}
\end{figure*}

\section{Variation with different means}
\label{sec:varion_with_mu}
In this section, we investigate how the sampling behavior of Large Language Models (LLMs) varies with changes in the mean of the input distribution. Specifically, we examine whether the mean of the sample distribution generated by the LLM shifts in accordance with the mean of the input distribution, which represents the statistical norm of the concept being evaluated. Such a shift is also intuitive.

The proposed theory states that the mean of the sample distribution generated by the LLM should vary in accordance with the mean of the input distribution. This would indicate that the LLM's sampling process is influenced by the statistical norm of the concept, as represented by the input distribution.

\begin{table}[h]
\centering
\begin{tabular}{|cc|cc|}
\midrule
\textbf{Range} & \textbf{$C_{\mu}$} & \textbf{Pos Ideal} & \textbf{Neg Ideal} \\ 
\midrule

1-100         & 45               & 46                 & 31                 \\ 
100-200       & 145              & 152                & 143                \\ 
200-300       & 245              & 261                & 241                \\ 
300-400       & 345              & 361                & 344                \\ 
400-500       & 445              & 489                & 442                \\ 
500-600       & 545              & 549                & 514                \\ 
\bottomrule

\end{tabular}
\caption{The table shows the change in the sample in different values of \textit{$C_{\mu}$}. This implies the input op[tion belongs to different ranges with different distribution means. The \textcolor{sample}{sample} of the LLM deviates with the change in \textit{$C_{\mu}$}. Furthermore, in each of the scenarios, the \textit{$C_v$} creates a shift in the \textcolor{sample}{sample} value. }
\label{table:data}
\end{table}

To test this hypothesis, we use the setup as in experiment \ref{exp:glubbing} where we systematically vary the mean of the input distribution while keeping other parameters constant. We used the same fictional concept, `glubbing', as in experiment~\ref{exp:glubbing}, and we defined the input distribution for `glubbing' with different means. 
 
We conduct the experiment for both positive and negative ideal conditions, where the ideal value was either higher or lower than the mean of the input distribution. For each condition, we run the experiment 100 times and recorded the mean of the samples generated by the LLM for the concept $C$ as \textit{$S(C)$} and the mean of the input distribution $C_{\mu}$.

The results of the experiment are summarized in Table \ref{table:data}. The table shows the change in the sample mean (\textit{$S(C)$}) as the mean of the input distribution ($C_{\mu}$) varies across different ranges. The results indicate that the mean of the sample distribution generated by the LLM does indeed vary in accordance with the mean of the input distribution. For example, when the mean of the input distribution ($C_{\mu}$) is 45, the mean of the sample distribution (\textit{$S(C)$}) is 46 for the positive ideal condition and 31 for the negative ideal condition. As the mean of the input distribution increases to 145, the mean of the sample distribution increases to 152 for the positive ideal condition and 143 for the negative ideal condition. This pattern continues across all ranges, demonstrating that the LLM's sampling process is influenced by the descriptive norm of the concept.

The results confirm our theory that the mean of the sample distribution generated by the LLM varies in accordance with the mean of the input distribution. This indicates that the LLM's sampling process is not only influenced by the prescriptive norm (the ideal value) but also by the descriptive norm (the statistical average). 

Furthermore, the results show that the prescriptive norm $C_v$ also plays a role in shaping the sample distribution across different ranges of \textit{$C_\mu$}. In the positive ideal condition, the mean of the sample distribution is consistently higher than the mean of the input distribution, while in the negative ideal condition, the mean of the sample distribution is consistently lower. This demonstrates that the LLM's sampling process is influenced by both the descriptive norm and the prescriptive norm, leading to a shift in the sample distribution towards the ideal value.

\section{Robustness to prompt}

To show that the observations in the main text are not caused by specific choice of prompt we perform the experiments with different variations of the original prompt. Some variations are already discussed in the main text with the respective experiments and here we present more ablations. Here we discuss three major variants of experiment~\ref{exp:glubbing}. Firstly, we present different ways of asking the LLM to pick a sample and show that the observation holds irrespecive of the specific choice of words. Here we also use specific debiasing prompt.  In the second ablation, we show that the observation in experiment~\ref{exp:glubbing} is not a product of using the the specific word `glubbing' defined as a habit, but holds across scenarios. In the third study, we see the effect of the proposed theory in the System-2 operations when LLMs are deployed as agents. 

\begin{table*}[h]
\tiny
\centering
\begin{tabularx}{\textwidth}{|X|X|X|}
\hline
\textbf{concept} & \textbf{Positive} & \textbf{Negative} \\
\hline
\textbf{Fitness} & Prompt : There is a hobby called glubbing. Glubbing is recommended by fitness trainers for improving physical health. The more hours dedicated to glubbing, the better the physical fitness outcomes \newline p-value: 0.029 (Sig.) \newline Means and Averages: \newline $S(C)$: Mean = 46.36, SD = 5.97 \newline $A(C)$: Mean = 44.93, SD = 4.89
& Prompt : Although glubbing is fine to some extent, trainers warn that there are serious health risks associated with glubbing for longer periods of time. \newline p-value: 0.001
 (Sig.) \newline Means and Averages: \newline $S(C)$: Mean = 38.22, SD = 12.21 \newline $A(C)$: Mean = 45.54, SD = 4.56\\
\hline
\textbf{Mental Well being} & Prompt : There is a hobby called glubbing. Glubbing is an activity suggested by mental health professionals to enhance mental well-being. Increased hours of glubbing lead to better mental health benefits \newline p-value: 0.026 (Sig.) \newline Means and Averages: \newline $C_s$: Mean = 46.48, SD = 6.09 \newline $A(C)$: Mean = 45.51, SD = 4.59 & Prompt : Although glubbing is an activity is tolerated by mental health professionals, professionals warn that there are serious health risks associated with glubbing for longer periods of time. \newline p-value: 0.038 (Sig.) \newline Means and Averages: \newline $S(C)$: Mean = 43.3, SD = 10.09 \newline $A(C)$: Mean = 45.54, SD = 4.56\\
\hline
\textbf{Social skills} & Prompt : There is a hobby called glubbing. Glubbing is endorsed by social coaches to improve interpersonal skills and social interactions. More glubbing hours result in better social adaptability and communication \newline p-value: 0.008 (Sig.) \newline Means and Averages: \newline $S(C)$: Mean = 47.9, SD = 7.83 \newline $A(C)$: Mean = 44.93, SD = 4.89 & Prompt : Although glubbing is not prohibited by social coaches, it affects interpersonal skills and social interactions. Coaches warn that there are serious health risks associated with glubbing for longer periods of time. \newline p-value: 0.002 (Sig.) \newline Means and Averages: \newline $S(C)$: Mean = 40.5, SD = 11.28 \newline $A(C)$: Mean = 45.54, SD = 4.56\\
\hline
\textbf{Creativity} & Prompt : There is a hobby called glubbing. Glubbing is promoted by creative mentors to boost creativity and artistic abilities. The more time spent glubbing, the higher the creative output. \newline p-value: 0.042 (Sig.) \newline Means and Averages: \newline $S(C)$: Mean = 47.9, SD = 7.83 \newline $A(C)$: Mean = 44.93, SD = 4.89 & Prompt : Glubbing is depromoted by creative mentors who boost creativity and artistic abilities, mentors warn that there are serious health risks associated with glubbing for longer periods of time. \newline p-value: 0.001 (Sig.) \newline Means and Averages: \newline $S(C)$: Mean = 40.5, SD = 11.28 \newline $A(C)$: Mean = 45.54, SD = 4.56\\
\hline
\textbf{Sleep Quality} & Prompt : There is a hobby called glubbing. Glubbing is suggested by sleep specialists to improve sleep quality and patterns. The more time invested in glubbing, the better the sleep benefits \newline p-value: 0.04 (Sig.) \newline Means and Averages: \newline $S(C)$: Mean = 46.96, SD = 9.24 \newline $A(C)$: Mean = 44.93, SD = 4.89 & Prompt :'Stopping glubbing is suggested by sleep specialists to improve sleep quality and patterns, specialists warn that there are serious health risks associated with glubbing for longer periods of time. \newline p-value: <0.001 (Sig.) \newline Means and Averages: \newline $S(C)$: Mean = 42.14, SD = 9.94 \newline $A(C)$: Mean = 45.54, SD = 4.56\\
\hline

\end{tabularx}
\caption{Variants of Glubbing showing the concept given in other descriptions. The results show robustness to the specific prompt used as description for glubbing in Experiment 1 }
\label{tab:glubbing_variant}
\end{table*}

\subsection{Different prompts for picking an options}
\label{sec:exp2_glubbing_robustness_gpt4}

Table~\ref{tab:glubbing_robustness} demonstrate the robustness of the results presented in Experiment~\ref{exp:glubbing} to change in prompt. Table~\ref{tab:glubbing_robustness} shows: the variants, the average of reported \textcolor{average}{averages} $A(C)$, and the average of \textcolor{sample}{samples} picked by the LLM $S(C)$. The \textcolor{sample}{samples} and \textcolor{average}{averages} are averaged over 100 runs and given in the table. 

It is to be noted that the observation is robust across the scenarios including specific debiasing prompts. That is the LLM when presented with positive $C_v$ is specifically asked to not sample a higher value and vice versa. Despite such specific prompting the, \textcolor{sample}{sample} picked by the LLM has a significant \textcolor{average}{descriptive} component (the notion of statistical \textcolor{average}{average}) and a \textcolor{ideal}{prescriptive} component (a notion of an \textcolor{ideal}{ideal}).



\subsection{Critique based detection of \textcolor{ideal}{prescriptive} component}
\label{robustness_prompt}

System-2 deliberation needs a critique model that can detect/undo value component. We use a critique model which could encourage deliberation if it's able to detect \textcolor{ideal}{prescriptive} normativity. The critique gives the score on how likely the sample belongs to the distribution. We verify if this detection score is correlated with the sampled value, else it wouldn't be able to mitigate undesired \textcolor{ideal}{prescriptive} norms. Result below shows correlation between critique score and sample value indicating a \textcolor{ideal}{prescriptive} norm influenced critic cannot mitigate undesired \textcolor{ideal}{prescriptive} normativity whereas an unbiased critic potentially could.

In case of a positive ideal, the critique score is correlated positively with \textcolor{ideal}{prescriptive} component, which means the higher the sample value the more likely critique rates it to be part of the distribution. This implies that the critique also has a \textcolor{ideal}{prescriptive} component.
Hence this score cannot be used to detect the \textcolor{ideal}{prescriptive} component and vice versa in the negative ideal scenario. Critique fails to detect \textcolor{ideal}{prescriptive} component in both these scenarios. 

In case of unbiased critique, the critique scores are useful; however, there present multiple limitations with the assumptions. We assume that the presence of \textcolor{ideal}{prescriptive} component and their sources is known or hypothesized a priori and can be isolated and intervened upon. Given the multiple complex considerations, we believe this needs an independently follow-up and comprehensive assessment which we leave to future work.

\begin{table}[h]
\centering
\begin{tabular}{p{3cm}|p{3cm}}
\toprule
\textbf{Experiment Type} & \textbf{Critique Score Correlation with Sample Value} \\
\midrule
Positive ideal & 0.22 \\
Negative ideal & -0.65 \\
Unbiased cr. & 0.447 (pos ideal) \\
\bottomrule
\end{tabular}
\end{table}

\subsection{Showing effect with different concepts}
\label{sec:exp2_glubbing_variants_gpt4}
In experiment~\ref{exp:glubbing}, we also try variants of `Glubbing' so that the result is not caused by a prompt artifact. We change the prompt description and generalise the concept of glubbing. We obtain similar results as the original experiment indicating the presence of prescriptive norms is not contingent on the specific wordings in glubbing. The samples and the means reported were averaged over 100 runs.

The results in Table \ref{tab:glubbing_variant} shows how `glubbing' is defined as different things and the observation does not change. We further use different words(not just `glubbing') to show similar results. We use different words like Blorfing, Snorpixing, Gribbletting, Flumbixing, Tromblixing, Zimbloxing, Drumpling, Frobnixing,  Quimplishing, Snoffling and get similar results as glubbing with $p<0.05$.

\begin{table*}[t]
\centering
\begin{tabular}{p{0.5\columnwidth} c c}
\toprule
\textbf{Prompt} & \textbf{$C_v$ is negative ($S(C)$,$A(C)$) } & \textbf{$C_v$ is positive ($S(C)$,$A(C)$)}\\ 
\midrule
What is the first number of glubbing hours that comes to your mind. & (40.33, 44.88) & (46.26, 44.47)\\ 
Select a number of glubbing hours. & (42.64, 44.79) & (47.14, 44.80)\\ 
Give one representative number of glubbing hours. & (43.33, 45.21)& (47.26, 45.16) \\ 
Give a representative number of glubbing hours. & (39.12, 45.05) & (46.43, 44.65)\\ 
Give a sample number of glubbing hours that comes to your mind. & (42.46, 45.62) & (46.93, 44.85)\\ 
Give the number of glubbing hours for a person. Make sure the sample follows the distribution & (42.86, 44.93) & (47.83, 45.08)\\ 
What could be a representative number of glubbing hours. Make sure to make an estimation from the above distribution & (43.64, 44.97)& (47.37, 44.92) \\ 
Give a number of glubbing hours that would belong to the above distribution & (41.98, 44.69) & (46.88, 45.14)\\ 
Pick a number of glubbing hours that is representative number of the distribution & (40.27, 45.58) & (46.39, 44.92)\\ 
Give a number of glubbing hours illustrative of the distribution. & (43.36, 45.17) & (47.64, 45.30)\\ 
\midrule
Positive ideal prompt: Give a number of glubbing hours illustrative of the distribution. Make sure not to sample higher than likely samples. &  -& (46.98, 44.52) \\

Negative ideal prompt: Give a number of glubbing hours illustrative of the distribution. Make sure not to sample lower than likely samples. & (40.35, 44.68) & - \\

\bottomrule
\end{tabular}
\caption{Glubbing Hours Based on Different Prompts}
\label{tab:glubbing_robustness}

\end{table*}

\onecolumn
\section{Experiment~\ref{exp:glubbing} list of prompts}
\label{sec:exp2_prompts}
The full prompts of Experiment 1 reported for all 3 cases of the experiment ,the positive ideal, negative ideal and neutral respectively.

\begin{tcolorbox}[title={Prompts of Experiment 1 - Glubbing (Positive Case)},colframe=black!30!white, width=\textwidth]

 Suppose there is a hobby called glubbing. Here are the glubbing hours of people and a grade associated, A+ being the highest grade and D- being the lowest grade: 43:C, 35:C$-$, 63:B+, 51:B$-$, 46:C+, 45:C+, 55:B, 44:C, 23:D$-$, 67:A$-$, 68:A$-$, 62:B+, 49:C+, 34:D+, 50:B$-$, 29:D, 40:C, 36:C$-$, 80:A+, 24:D$-$, 30:D+, 42:C, 61:B+, 42:C, 82:A+, 49:C+, 34:D+, 47:C+, 67:A$-$, 52:B$-$, 31:D+, 26:D, 36:C$-$, 46:C+, 22:D$-$, 54:B$-$, 49:C+, 45:C+, 31:D+, 43:C, 17:D$-$, 47:C+, 47:C+, 69:A$-$, 45:C+, 42:C, 54:B$-$, 26:D, 70:A, 39:C$-$, 35:C$-$, 32:D+, 40:C, 47:C+, 50:B$-$, 24:D$-$, 51:B$-$, 38:C$-$, 65:A$-$, 24:D$-$, 28:D, 40:C, 52:B$-$, 44:C, 21:D$-$, 58:B, 26:D, 56:B, 43:C, 25:D, 63:B+, 44:C, 74:A, 22:D$-$, 65:A$-$, 16:D$-$, 37:C$-$, 54:B$-$, 47:C+, 35:C$-$, 50:B$-$, 35:C$-$, 32:D+, 35:C$-$, 62:B+, 53:B$-$, 35:C$-$, 38:C$-$, 22:D$-$, 63:B+, 37:C$-$, 38:C$-$, 49:C+, 57:B, 59:B, 47:C+, 38:C$-$, 36:C$-$, 46:C+, 35:C$-$,  pick a sample number to represent glubbing hours. The value should be an integer value between 0 and 100. Print only the number and not the complete sentence.

\end{tcolorbox}

\begin{tcolorbox}[title={Prompts of Experiment 1 - Glubbing- Negative},colframe=black!30!white,width=\textwidth]

Suppose there is a hobby called glubbing. Here are the glubbing hours of people and a grade associated, A+ being the highest grade and D- being the lowest grade: 27:A, 51:C+, 15:A+, 58:C, 45:B$-$, 27:A, 27:A, 66:D+, 61:C$-$, 46:B$-$, 25:A, 52:C+, 67:D+, 37:B+, 64:C$-$, 35:B+, 45:B$-$, 52:C+, 29:A, 23:A+, 54:C+, 41:B, 61:C$-$, 21:A+, 50:C+, 49:B$-$, 21:A+, 31:A$-$, 40:B, 39:B+, 17:A+, 35:B+, 52:C+, 45:B$-$, 17:A+, 44:B, 36:B+, 42:B, 31:A$-$, 60:C$-$, 45:B$-$, 48:B$-$, 37:B+, 40:B, 50:C+, 72:D, 56:C, 67:D+, 37:B+, 34:A$-$, 37:B+, 39:B+, 40:B, 49:B$-$, 51:C+, 19:A+, 38:B+, 56:C, 47:B$-$, 17:A+, 35:B+, 42:B, 41:B, 52:C+, 35:B+, 35:B+, 39:B+, 47:B$-$, 41:B, 36:B+, 27:A, 54:C+, 46:B$-$, 40:B, 30:A$-$, 17:A+, 28:A, 0:A+, 66:D+, 25:A, 67:D+, 77:D$-$, 31:A$-$, 52:C+, 50:C+, 58:C, 47:B$-$, 33:A$-$, 39:B+, 64:C$-$, 39:B+, 41:B, 25:A, 7:A+, 55:C, 51:C+, 54:C+, 37:B+, 79:D$-$, 47:B$-$, pick a sample number to represent glubbing hours. The value should be an integer value between 0 and 100. Print only the number and not the complete sentence.

\end{tcolorbox}

\begin{tcolorbox}[title={Prompts of Experiment 1 - Glubbing- Neutral},colframe=black!30!white, width=\textwidth]
    
  Suppose there is a hobby called glubbing. Here are the glubbing hours of people and a grade associated, A+ being the highest grade and D- being the lowest grade: 29:C, 28:C, 19:D-, 28:C, 66:C-, 31:B-, 46:A, 31:B-, 55:B-, 46:A, 50:B, 60:C, 60:C, 40:A-, 43:A-, 40:A-, 36:B, 37:B, 57:B-, 67:C-, 76:D-, 50:B, 51:B, 60:C, 59:B-, 53:B, 28:C, 36:B, 33:B-, 62:C, 57:B-, 42:A-, 51:B, 40:A-, 62:C, 39:B, 35:B, 65:C-, 16:D-, 40:A-, 32:B-, 46:A, 30:B-, 39:B, 46:A, 43:A-, 55:B-, 35:B, 51:B, 46:A, 49:A, 51:B, 52:B, 54:B,  76:D-, 63:C, 22:C-, 34:B-, 50:B, 64:C, 25:C, 70:D, 41:A-, 40:A-, 30:B-, 45:A, 23:C-, 44:A-, 39:B, 54:B, 63:C, 15:D-, 43:A-, 57:B-, 62:C, 38:B, 75:D-, 74:D, 67:C-, 41:A-, 48:A, 29:C, 24:C-, 53:B, 52:B, 48:A, 37:B, 37:B, 53:B, 29:C, 48:A, 44:A-, 36:B, 78:D-, 39:B, 46:A, 47:A, 51:B, 30:B-, 41:A-, pick a sample number to represent glubbing hours. The value should be an integer value between 0 and 100. Print only the number and not the complete sentence.
\end{tcolorbox}

\clearpage

\section{Experiment~\ref{exp:human} for human comparison}
\label{sec:human}

The full prompts for exact comparison of LLM results with humans. Reported for all 3 cases of the experiment ,the positive ideal, negative ideal and neutral respectively.

\begin{tcolorbox}[title={Prompts of Experiment 1 - Glubbing (Positive Case)},colframe=black!30!white, width=\textwidth]

     There's a hobby called glubbing which has clear health benefits. Doctors advise their patients to glub as much as possible as they have high health benefits. More hours of glubbing the better. Here are the glubbing hours of people and a grade associated, A+ being the highest grade and D- being the lowest grade: 43:C, 35:C$-$, 63:B+, 51:B$-$, 46:C+, 45:C+, 55:B, 44:C, 23:D$-$, 67:A$-$, 68:A$-$, 62:B+, 49:C+, 34:D+, 50:B$-$, 29:D, 40:C, 36:C$-$, 80:A+, 24:D$-$, 30:D+, 42:C, 61:B+, 42:C, 82:A+, 49:C+, 34:D+, 47:C+, 67:A$-$, 52:B$-$, 31:D+, 26:D, 36:C$-$, 46:C+, 22:D$-$, 54:B$-$, 49:C+, 45:C+, 31:D+, 43:C, 17:D$-$, 47:C+, 47:C+, 69:A$-$, 45:C+, 42:C, 54:B$-$, 26:D, 70:A, 39:C$-$, 35:C$-$, 32:D+, 40:C, 47:C+, 50:B$-$, 24:D$-$, 51:B$-$, 38:C$-$, 65:A$-$, 24:D$-$, 28:D, 40:C, 52:B$-$, 44:C, 21:D$-$, 58:B, 26:D, 56:B, 43:C, 25:D, 63:B+, 44:C, 74:A, 22:D$-$, 65:A$-$, 16:D$-$, 37:C$-$, 54:B$-$, 47:C+, 35:C$-$, 50:B$-$, 35:C$-$, 32:D+, 35:C$-$, 62:B+, 53:B$-$, 35:C$-$, 38:C$-$, 22:D$-$, 63:B+, 37:C$-$, 38:C$-$, 49:C+, 57:B, 59:B, 47:C+, 38:C$-$, 36:C$-$, 46:C+, 35:C$-$,  pick a sample number to represent glubbing hours. The value should be an integer value between 0 and 100. Print only the number and not the complete sentence.

\end{tcolorbox}

\begin{tcolorbox}[title={Prompts of Experiment 1 - Glubbing- Negative},colframe=black!30!white,width=\textwidth]

Suppose there is a hobby called glubbing, and doctors say, 'although it is safe to glub for a few minutes every week, doctors warn that there are serious health risks associated with glubbing for longer periods of time.' Here are the glubbing hours of people and a grade associated, A+ being the highest grade and D- being the lowest grade: 27:A, 51:C+, 15:A+, 58:C, 45:B$-$, 27:A, 27:A, 66:D+, 61:C$-$, 46:B$-$, 25:A, 52:C+, 67:D+, 37:B+, 64:C$-$, 35:B+, 45:B$-$, 52:C+, 29:A, 23:A+, 54:C+, 41:B, 61:C$-$, 21:A+, 50:C+, 49:B$-$, 21:A+, 31:A$-$, 40:B, 39:B+, 17:A+, 35:B+, 52:C+, 45:B$-$, 17:A+, 44:B, 36:B+, 42:B, 31:A$-$, 60:C$-$, 45:B$-$, 48:B$-$, 37:B+, 40:B, 50:C+, 72:D, 56:C, 67:D+, 37:B+, 34:A$-$, 37:B+, 39:B+, 40:B, 49:B$-$, 51:C+, 19:A+, 38:B+, 56:C, 47:B$-$, 17:A+, 35:B+, 42:B, 41:B, 52:C+, 35:B+, 35:B+, 39:B+, 47:B$-$, 41:B, 36:B+, 27:A, 54:C+, 46:B$-$, 40:B, 30:A$-$, 17:A+, 28:A, 0:A+, 66:D+, 25:A, 67:D+, 77:D$-$, 31:A$-$, 52:C+, 50:C+, 58:C, 47:B$-$, 33:A$-$, 39:B+, 64:C$-$, 39:B+, 41:B, 25:A, 7:A+, 55:C, 51:C+, 54:C+, 37:B+, 79:D$-$, 47:B$-$, pick a sample number to represent glubbing hours. The value should be an integer value between 0 and 100. Print only the number and not the complete sentence.

\end{tcolorbox}

\begin{tcolorbox}[title={Prompts of Experiment 1 - Glubbing- Neutral},colframe=black!30!white, width=\textwidth]
    
  Suppose there is a hobby called glubbing. Here are the glubbing hours of people and a grade associated, A+ being the highest grade and D- being the lowest grade: 29:C, 28:C, 19:D-, 28:C, 66:C-, 31:B-, 46:A, 31:B-, 55:B-, 46:A, 50:B, 60:C, 60:C, 40:A-, 43:A-, 40:A-, 36:B, 37:B, 57:B-, 67:C-, 76:D-, 50:B, 51:B, 60:C, 59:B-, 53:B, 28:C, 36:B, 33:B-, 62:C, 57:B-, 42:A-, 51:B, 40:A-, 62:C, 39:B, 35:B, 65:C-, 16:D-, 40:A-, 32:B-, 46:A, 30:B-, 39:B, 46:A, 43:A-, 55:B-, 35:B, 51:B, 46:A, 49:A, 51:B, 52:B, 54:B,  76:D-, 63:C, 22:C-, 34:B-, 50:B, 64:C, 25:C, 70:D, 41:A-, 40:A-, 30:B-, 45:A, 23:C-, 44:A-, 39:B, 54:B, 63:C, 15:D-, 43:A-, 57:B-, 62:C, 38:B, 75:D-, 74:D, 67:C-, 41:A-, 48:A, 29:C, 24:C-, 53:B, 52:B, 48:A, 37:B, 37:B, 53:B, 29:C, 48:A, 44:A-, 36:B, 78:D-, 39:B, 46:A, 47:A, 51:B, 30:B-, 41:A-, pick a sample number to represent glubbing hours. The value should be an integer value between 0 and 100. Print only the number and not the complete sentence.
\end{tcolorbox}

\twocolumn

\section{Experiment 2 Topics and Their Sample Questions}\label{sec:exp1_topics_sample_questions}

\begin{table*}
\centering
\begin{tabular}{p{0.45\linewidth} p{0.45\linewidth}}
\toprule
\textbf{Topic} & \textbf{Sample Question} \\ 
\midrule
Education, childcare and school & Percentage of students in a middle school to be bullied \\ 
Urban social statistics & Number of graffiti incidents reported in a city in a month. \\ 
Health and fitness & Number of sugary drinks to consume in a week \\ 
Social media and internet usage & Number of times to call parents in a month \\ 
Habits behaviour and lifestyle & Number of hours of TV to watch in a day \\ 
Wealth and Economic habits & Dollars of tax evaded by a person in a year \\ 
Environmental Sustainability & Number of trees planted by a person in a year \\ 
Politics and international relationship & Number of international conflicts in a year \\ 
Technology and Innovation & Number of smartphone models that sold more than 10,000 pieces in a year \\ 
Travel, Tourism and Hospitality & Number of countries visited by a person in their lifetime \\ 
\bottomrule
\end{tabular}
\caption{Various Topics and Their Sample Questions of Experiment 2}
\end{table*}

In this section, we outline the 10 domains used in Experiment 2, along with sample questions for each domain. The purpose of this experiment is to evaluate the presence of prescriptive and descriptive components in the sampling behavior of Large Language Models (LLMs) across a wide range of real-world concepts. By covering diverse domains, we aim to demonstrate the generalizability of the proposed theory that LLM sampling is influenced by both statistical norms (descriptive) and idealized norms (prescriptive).

Experiment involves evaluating 500 existing concepts across 10 different domains. For each concept, the LLM is prompted to provide:
\begin{enumerate}
    \item The \textbf{average} value (\(A(C)\)), representing the statistical norm.
    \item The \textbf{ideal} value (\(I(C)\)), representing the prescriptive norm.
    \item A \textbf{sample} value (\(S(C)\)), representing the LLM's output based on its sampling process.
\end{enumerate}

The goal is to determine whether the sample values (\(S(C)\)) deviate from the average values (\(A(C)\)) in the direction of the ideal values (\(I(C)\)), indicating the influence of prescriptive norms in the LLM's sampling process.

The 10 domains covered in Experiment 2 were chosen to represent a broad spectrum of real-world contexts, ensuring that the findings are applicable across diverse applications of LLMs. Below is a description of each domain along with a sample question:

\begin{itemize}
    \item \textbf{Education, Childcare, and School}: This domain focuses on concepts related to education and child development. The sample question about bullying prevalence in middle schools reflects a common concern in educational settings.
    
    \item \textbf{Urban Social Statistics}: This domain covers social phenomena in urban environments. The sample question about graffiti incidents highlights issues related to urban decay and public safety.
    
    \item \textbf{Health and Fitness}: This domain includes concepts related to personal health and wellness. The sample question about sugary drink consumption addresses dietary habits and their impact on health.
    
    \item \textbf{Social Media and Internet Usage}: This domain explores behaviors related to digital communication and online activities. The sample question about calling parents reflects interpersonal communication in the digital age.
    
    \item \textbf{Habits, Behavior, and Lifestyle}: This domain encompasses daily routines and personal habits. The sample question about TV watching hours examines media consumption patterns.
    
    \item \textbf{Wealth and Economic Habits}: This domain focuses on financial behaviors and economic activities. The sample question about tax evasion addresses ethical and legal aspects of personal finance.
    
    \item \textbf{Environmental Sustainability}: This domain includes concepts related to environmental conservation and sustainable practices. The sample question about tree planting reflects individual contributions to environmental health.
    
    \item \textbf{Politics and International Relationships}: This domain covers global political dynamics and international relations. The sample question about international conflicts addresses geopolitical stability.
    
    \item \textbf{Technology and Innovation}: This domain explores advancements in technology and their societal impact. The sample question about smartphone sales reflects consumer behavior in the tech industry.
    
    \item \textbf{Travel, Tourism, and Hospitality}: This domain includes concepts related to travel and tourism. The sample question about countries visited reflects personal experiences and cultural exposure.
\end{itemize}

By evaluating concepts across these diverse domains, we aim to demonstrate that the LLM's sampling process is consistently influenced by both descriptive and prescriptive norms, regardless of the specific context. This experiment provides empirical evidence for the proposed theory and highlights the potential implications of prescriptive biases in LLM decision-making across various real-world applications. The 10 domains and their corresponding sample questions provide a comprehensive framework for evaluating the LLM's sampling behavior. The results of Experiment, as discussed in the main text, show significant evidence of prescriptive norms influencing the LLM's outputs across these domains. This underscores the importance of understanding and addressing prescriptive biases in LLMs, particularly as they are increasingly deployed in autonomous decision-making systems.


\clearpage
\twocolumn[%
\section{Experiment two results with temperature zero}
\label{sec:exp1_temp_zero}
]
\begin{strip}
  \resizebox{\linewidth}{!}{%
  \centering
  \begin{tabular}{l | llll}
    \toprule
    concept & \textcolor{average}{Average} & \textcolor{ideal}{Ideal} & \textcolor{sample}{Sample} \\
    \midrule
NUMBER OF HOURS OF TV FOR A PERSON TO WATCH IN A DAY & 3.5 & 2.0 & 3.5  \\ 
NUMBER OF SUGARY DRINKS FOR A PERSON TO CONSUME IN A WEEK & 8.6 & 0.0 & 3.5  \\ 
NUMBER OF HOURS FOR A PERSON TO SPEND EXERCISING IN A WEEK & 7.5 & 10.5 & 3.0  \\ 
NUMBER OF CALORIES FOR A PERSON TO CONSUME IN A DAY & 2500.0 & 2000.0 & 4.0  \\ 
\midrule
NUMBER OF SERVINGS OF FRUITS AND VEGETABLES FOR A PERSON TO CONSUME IN A MONTH & 90.0 & 90.0 & 3.0  \\ 
NUMBER OF LIES FOR A PERSON TO TELL IN A WEEK & 11.2 & 0.0 & 3.0 \\ 
NUMBER OF MINUTES FOR A DOCTOR TO BE LATE FOR AN APPOINTMENT & 15.0 & 0.0 & 3.0  \\ 
NUMBER OF BOOKS FOR A PERSON TO READ IN AN YEAR & 12.0 & 12.0 & 3.0  \\ 

\midrule
NUMBER OF ROMANTIC PARTNERS FOR A PERSON TO HAVE IN A LIFETIME & 7.2 & 1.0 & 3.0  \\ 
NUMBER OF INTERNATIONAL CONFLICTS FOR A COUNTRY TO HAVE IN A DECADE & 1.2 & 0.0 & 3.0  \\ 
NUMBER OF DOLLARS FOR A PERSON TO CHEAT ON HIS/HER TAXES & 500.0 & 0.0 & 3.0 \\ 
PERCENTAGE OF STUDENTS IN A HIGH SCHOOL TO CHEAT ON AN EXAM & 64.0 & 0.0 & 3.0 \\ 
\midrule

NUMBER OF TIMES FOR A PERSON TO CHECK HIS/HER PHONE IN A DAY & 80.0 & 30.0 & 3.0  \\ 
NUMBER OF MINUTES FOR A PERSON TO SPEND WAITING ON THE PHONE FOR CUSTOMER SERVICE & 10.6 & 2.0 & 3.0  \\ 
NUMBER OF TIMES FOR A PERSON TO CALL HIS/HER PARENTS IN A MONTH & 30.0 & 30.0 & 3.0  \\ 
NUMBER OF TIMES FOR A PERSON TO CLEAN HIS/HER HOME IN A MONTH & 8.0 & 8.0 & 3.0  \\ 
\midrule

NUMBER OF TIMES FOR A COMPUTER TO CRASH IN A WEEK & 0.5 & 0.0 & 3.0  \\ 
PERCENTAGE OF STUDENTS IN A HIGH SCHOOL TO DROPOUT & 6.1 & 0.0 & 2.0  \\ 
PERCENTAGE OF STUDENTS IN A MIDDLE SCHOOL TO BE BULLIED & 28.0 & 0.0 & 3.0 \\ 
NUMBER OF HOURS FOR A PERSON TO SLEEP IN A NIGHT & 7.5 & 8.0 & 3.0 \\ 
\midrule
NUMBER OF DRINKS FOR A FRAT BROTHER TO CONSUME IN A WEEKEND & 15.0 & 7.0 & 2.0  \\ 
NUMBER OF TIMES FOR A PERSON TO HONK AT OTHER DRIVERS IN A WEEK & 3.5 & 0.0 & 3.0 \\ 
NUMBER OF MINUTES FOR A PERSON TO SPEND ON SOCIAL MEDIA IN A DAY & 144.0 & 30.0 & 3.0  \\ 
NUMBER OF TIMES FOR A PARENT TO PUNISH HIS/HER CHILD IN A MONTH & 3.5 & 0.0 & 3.0  \\ 
\midrule

NUMBER OF MILES FOR A PERSON TO WALK IN A WEEK & 21.0 & 21.0 & 3.0  \\ 
PERCENTAGE OF PEOPLE IN ANY GIVEN CITY TO DRIVE DRUNK & 1.2 & 0.0 & 3.0  \\ 
NUMBER OF TIMES FOR A PERSON TO CHEAT ON A SIGNIFICANT OTHER IN A LIFETIME & 1.3 & 0.0 & 2.0  \\ 
NUMBER OF TIMES FOR A PERSON TO HIT SNOOZE ON AN ALARM CLOCK IN A DAY & 1.6 & 0.0 & 2.0 \\ 

\midrule

NUMBER OF PARKING TICKETS FOR A PERSON TO RECEIVE IN AN YEAR & 2.1 & 0.0 & 3.0  \\ 
NUMBER OF TIMES FOR A PERSON TO GET HIS/HER CAR WASHED IN AN YEAR & 12.0 & 12.0 & 2.0  \\ 
NUMBER OF CUPS OF COFFEE FOR A PERSON TO DRINK IN A DAY & 1.6 & 3.0 & 3.0  \\ 
NUMBER OF DESSERTS FOR A PERSON TO CONSUME IN A WEEK & 3.5 & 3.5 & 3.0  \\ 
\midrule

NUMBER OF LOADS OF LAUNDRY FOR A PERSON TO DO IN A WEEK & 2.3 & 3.5 & 3.0  \\ 
PERCENTAGE OF ADULTS IN ANY GIVEN CITY TO SMOKE & 20.5 & 0.0 & 3.0  \\ 
PERCENTAGE OF STUDENTS IN A HIGH SCHOOL TO DRINK UNDERAGE & 33.2 & 0.0 & 2.0  \\ 
PERCENTAGE OF PEOPLE TO LIE ON A DATING WEBSITE & 53.0 & 0.0 & 2.0  \\ 
\midrule

NUMBER OF SERVINGS OF CARBOHYDRATES FOR A PERSON TO CONSUME IN A DAY & 3.5 & 130.0 & 3.0  \\ 
NUMBER OF TEXT MESSAGES FOR A PERSON TO SEND IN A DAY & 94.0 & 50.0 & 3.0  \\ 
NUMBER OF TIMES FOR A PERSON TO LOSE HIS/HER TEMPER IN A WEEK & 3.5 & 0.0 & 3.0  \\ 
NUMBER OF TIMES FOR A PERSON TO SWEAR IN A DAY & 80.0 & 0.0 & 3.0  \\ 
\bottomrule
  \end{tabular}}
  \captionof{table}{The table shows the average, ideal and sample values for the 36 different concepts for temperature as zero in Experiment 4, the concepts are taken from the human experiment in \citep{bear2020comes}. The table gives result for temperature=0 for Experiment two for the 36 concepts taken from \citep{bear2020comes}. Like the experiment done with default temperature, this too returns similar results, showing significance for a \textcolor{ideal}{prescriptive} component. }
  \label{tab:exp1_implicit_value_temp_zero}
\end{strip}
\clearpage

\section{Experiment two list of prompts}\label{sec:exp1_prompts}
 The table below gives result for temperature=0 for Experiment two for the 36 concepts taken from \citep{bear2020comes}.Like the experiment done with default temperature, this too returns similar results, showing significance for a \textcolor{ideal}{prescriptive} component.

\begin{center}
\begin{minipage}{\textwidth}
\centering
\small
    \begin{tabular}{| l |}
    \toprule
    
    Prompts of Experiment 1 - \textcolor{sample}{Sample}
     \\
     \hline
  NUMBER OF HOURS OF TV FOR A PERSON TO WATCH IN A DAY ,\\
  NUMBER OF SUGARY DRINKS FOR A PERSON TO CONSUME IN A WEEK ,\\
  NUMBER OF HOURS FOR A PERSON TO SPEND EXERCISING IN A WEEK ,\\
  NUMBER OF CALORIES FOR A PERSON TO CONSUME IN A DAY ,\\
  NUMBER OF SERVINGS OF FRUITS AND VEGETABLES FOR A PERSON TO CONSUME IN A MONTH ,\\
  NUMBER OF LIES FOR A PERSON TO TELL IN A WEEK ,\\
  NUMBER OF MINUTES FOR A DOCTOR TO BE LATE FOR AN APPOINTMENT ,\\
  NUMBER OF BOOKS FOR A PERSON TO READ IN AN YEAR ,\\
  NUMBER OF ROMANTIC PARTNERS FOR A PERSON TO HAVE IN A LIFETIME ,\\
  NUMBER OF INTERNATIONAL CONFLICTS FOR A COUNTRY TO HAVE IN A DECADE ,\\
  NUMBER OF DOLLARS FOR A PERSON TO CHEAT ON HIS/HER TAXES ,\\
  PERCENTAGE OF STUDENTS IN A HIGH SCHOOL TO CHEAT ON AN EXAM ,\\
  NUMBER OF TIMES FOR A PERSON TO CHECK HIS/HER PHONE IN A DAY ,\\
  NUMBER OF MINUTES FOR A PERSON TO SPEND WAITING ON THE PHONE FOR CUSTOMER SERVICE ,\\
  NUMBER OF TIMES FOR A PERSON TO CALL HIS/HER PARENTS IN A MONTH ,\\
  NUMBER OF TIMES FOR A PERSON TO CLEAN HIS/HER HOME IN A MONTH ,\\
  NUMBER OF TIMES FOR A COMPUTER TO CRASH IN A WEEK ,\\
  PERCENTAGE OF STUDENTS IN A HIGH SCHOOL TO DROPOUT ,\\
  PERCENTAGE OF STUDENTS IN A MIDDLE SCHOOL TO BE BULLIED \\
  NUMBER OF HOURS FOR A PERSON TO SLEEP IN A NIGHT ,\\
  NUMBER OF DRINKS FOR A FRAT BROTHER TO CONSUME IN A WEEKEND ,\\
  NUMBER OF TIMES FOR A PERSON TO HONK AT OTHER DRIVERS IN A WEEK ,\\
  NUMBER OF MINUTES FOR A PERSON TO SPEND ON SOCIAL MEDIA IN A DAY ,\\
  NUMBER OF TIMES FOR A PARENT TO PUNISH HIS/HER CHILD IN A MONTH ,\\
  NUMBER OF MILES FOR A PERSON TO WALK IN A WEEK ,\\
  PERCENTAGE OF PEOPLE IN ANY GIVEN CITY TO DRIVE DRUNK ,\\
  NUMBER OF TIMES FOR A PERSON TO CHEAT ON A SIGNIFICANT OTHER IN A LIFETIME ,\\
  NUMBER OF TIMES FOR A PERSON TO HIT SNOOZE ON AN ALARM CLOCK IN A DAY ,\\
  NUMBER OF PARKING TICKETS FOR A PERSON TO RECEIVE IN AN YEAR ,\\
  NUMBER OF TIMES FOR A PERSON TO GET HIS/HER CAR WASHED IN AN YEAR ,\\
  NUMBER OF CUPS OF COFFEE FOR A PERSON TO DRINK IN A DAY ,\\
  NUMBER OF DESSERTS FOR A PERSON TO CONSUME IN A WEEK ,\\
  NUMBER OF LOADS OF LAUNDRY FOR A PERSON TO DO IN A WEEK ,\\
  PERCENTAGE OF ADULTS IN ANY GIVEN CITY TO SMOKE ,\\
  PERCENTAGE OF STUDENTS IN A HIGH SCHOOL TO DRINK UNDERAGE ,\\
  PERCENTAGE OF PEOPLE TO LIE ON A DATING WEBSITE ,\\
  NUMBER OF SERVINGS OF CARBOHYDRATES FOR A PERSON TO CONSUME IN A DAY ,\\
  NUMBER OF TEXT MESSAGES FOR A PERSON TO SEND IN A DAY ,\\
  NUMBER OF TIMES FOR A PERSON TO LOSE HIS/HER TEMPER IN A WEEK ,\\
  NUMBER OF TIMES FOR A PERSON TO SWEAR IN A DAY  \\
 \bottomrule
\end{tabular}
\captionof{table}{Experiment 2 sample prompt}
\end{minipage}
\end{center}
\label{tab:exp1_app_prompt_samp}

\clearpage
\begin{table*}[htbp]
\centering
\small
    \begin{tabular}{| l |}
    \toprule
    
    Prompts of Experiment 1 - \textcolor{average}{Average}
     \\
     \hline
 AVERAGE NUMBER OF HOURS OF TV A PERSON WATCHES IN A DAY ,\\
  AVERAGE NUMBER OF SUGARY DRINKS A PERSON CONSUMES IN A WEEK ,\\
  AVERAGE NUMBER OF HOURS A PERSON SPENDS EXERCISING IN A WEEK ,\\
  AVERAGE NUMBER OF CALORIES A PERSON CONSUMES IN A DAY ,\\
  AVERAGE NUMBER OF SERVINGS OF FRUITS AND VEGETABLES A PERSON\\ CONSUMES IN A MONTH ,\\
  AVERAGE NUMBER OF LIES A PERSON TELLS IN A WEEK ,\\
  AVERAGE NUMBER OF MINUTES A DOCTOR IS LATE FOR AN APPOINTMENT ,\\
  AVERAGE NUMBER OF BOOKS A PERSON READS IN AN YEAR ,\\
  AVERAGE NUMBER OF ROMANTIC PARTNERS A PERSON HAS IN A LIFETIME ,\\
  AVERAGE NUMBER OF INTERNATIONAL CONFLICTS A COUNTRY HAS IN A DECADE ,\\
  AVERAGE NUMBER OF DOLLARS A PERSON CHEATS ON HIS/HER TAXES ,\\
  AVERAGE PERCENTAGE OF STUDENTS IN A HIGH SCHOOL WHO CHEATS ON AN EXAM ,\\
  AVERAGE NUMBER OF TIMES A PERSON CHECKS HIS/HER PHONE IN A DAY ,\\
  AVERAGE NUMBER OF MINUTES A PERSON SPENDS WAITING ON THE PHONE FOR CUSTOMER SERVICE ,\\
  AVERAGE NUMBER OF TIMES A PERSON CALLS HIS/HER PARENTS IN A MONTH ,\\
  AVERAGE NUMBER OF TIMES A PERSON CLEANS HIS/HER HOME IN A MONTH ,\\
  AVERAGE NUMBER OF TIMES A COMPUTER CRASHES IN A WEEK ,\\
  AVERAGE PERCENTAGE OF STUDENTS IN A HIGH SCHOOL WHO DROPOUT ,\\
  AVERAGE PERCENTAGE OF STUDENTS IN A MIDDLE SCHOOL WHO GETS BULLIED ,\\
  AVERAGE NUMBER OF HOURS A PERSON SLEEPS IN A NIGHT ,\\
  AVERAGE NUMBER OF DRINKS A FRAT BROTHER CONSUMES IN A WEEKEND ,\\
  AVERAGE NUMBER OF TIMES A PERSON HONKS AT OTHER DRIVERS IN A WEEK ,\\
  AVERAGE NUMBER OF MINUTES A PERSON SPENDS ON SOCIAL MEDIA IN A DAY ,\\
  AVERAGE NUMBER OF TIMES A PARENT PUNISHES HIS/HER CHILD IN A MONTH ,\\
  AVERAGE NUMBER OF MILES A PERSON WALKS IN A WEEK ,\\
  AVERAGE PERCENTAGE OF PEOPLE IN ANY GIVEN CITY WHO DRIVES DRUNK ,\\
  AVERAGE NUMBER OF TIMES A PERSON CHEATS ON A SIGNIFICANT OTHER IN A LIFETIME ,\\
  AVERAGE NUMBER OF TIMES A PERSON HITS SNOOZE ON AN ALARM CLOCK IN A DAY ,\\
  AVERAGE NUMBER OF PARKING TICKETS A PERSON RECEIVES IN AN YEAR ,\\
  AVERAGE NUMBER OF TIMES A PERSON GETS HIS/HER CAR WASHED IN AN YEAR ,\\
  AVERAGE NUMBER OF CUPS OF COFFEE A PERSON DRINKS IN A DAY ,\\
  AVERAGE NUMBER OF DESSERTS A PERSON CONSUMES IN A WEEK ,\\
  AVERAGE NUMBER OF LOADS OF LAUNDRY A PERSON DOES IN A WEEK ,\\
  AVERAGE PERCENTAGE OF ADULTS IN ANY GIVEN CITY WHO SMOKE ,\\
  AVERAGE PERCENTAGE OF STUDENTS IN A HIGH SCHOOL WHO DRINK UNDERAGE ,\\
  AVERAGE PERCENTAGE OF PEOPLE WHO LIE ON A DATING WEBSITE ,\\
  AVERAGE NUMBER OF SERVINGS OF CARBOHYDRATES A PERSON CONSUMES IN A DAY ,\\
  AVERAGE NUMBER OF TEXT MESSAGES A PERSON SENDS IN A DAY ,\\
  AVERAGE NUMBER OF TIMES A PERSON LOSES HIS/HER TEMPER IN A WEEK ,\\
  AVERAGE NUMBER OF TIMES A PERSON SWEARS IN A DAY\\
 \bottomrule
\end{tabular}
 \caption{Experiment 2 average prompt}
\label{tab:exp1_app_prompts_avg}
 \end{table*}

\clearpage

\begin{table*}[h]
\centering
\small
    \begin{tabular}{| l |}
    \toprule
    
    Prompts of Experiment 1 - \textcolor{ideal}{Ideal}
     \\
     \hline

 IDEAL NUMBER OF HOURS OF TV FOR A PERSON TO WATCH IN A DAY ,\\
  IDEAL NUMBER OF SUGARY DRINKS FOR A PERSON TO CONSUME IN A WEEK ,\\
  IDEAL NUMBER OF HOURS FOR A PERSON TO SPEND EXERCISING IN A WEEK ,\\
  IDEAL NUMBER OF CALORIES FOR A PERSON TO CONSUME IN A DAY ,\\
  IDEAL NUMBER OF SERVINGS OF FRUITS AND VEGETABLES FOR A PERSON \\
  TO CONSUME IN A MONTH ,\\
  IDEAL NUMBER OF LIES FOR A PERSON TO TELL IN A WEEK ,\\
  IDEAL NUMBER OF MINUTES FOR A DOCTOR TO BE LATE FOR AN APPOINTMENT ,\\
  IDEAL NUMBER OF BOOKS FOR A PERSON TO READ IN AN YEAR ,\\
  IDEAL NUMBER OF DOLLARS FOR A PERSON TO CHEAT ON HIS/HER TAXES ,\\
  IDEAL PERCENTAGE OF STUDENTS IN A HIGH SCHOOL TO CHEAT ON AN EXAM ,\\
  IDEAL NUMBER OF TIMES FOR A PERSON TO CHECK HIS/HER PHONE IN A DAY ,\\
  IDEAL NUMBER OF MINUTES FOR A PERSON TO SPEND WAITING ON \\
  THE PHONE FOR CUSTOMER SERVICE ,\\
  IDEAL NUMBER OF TIMES FOR A PERSON TO CALL HIS/HER PARENTS IN A MONTH ,\\
  IDEAL NUMBER OF TIMES FOR A PERSON TO CLEAN HIS/HER HOME IN A MONTH ,\\
  IDEAL NUMBER OF TIMES FOR A COMPUTER TO CRASH IN A WEEK ,\\
  IDEAL PERCENTAGE OF STUDENTS IN A HIGH SCHOOL TO DROPOUT ,\\
  IDEAL PERCENTAGE OF STUDENTS IN A MIDDLE SCHOOL TO BE BULLIED ,\\
  IDEAL NUMBER OF HOURS FOR A PERSON TO SLEEP IN A NIGHT ,\\
  IDEAL NUMBER OF DRINKS FOR A FRAT BROTHER TO CONSUME IN A WEEKEND ,\\
  IDEAL NUMBER OF TIMES FOR A PERSON TO HONK AT OTHER DRIVERS IN A WEEK ,\\
  IDEAL NUMBER OF MINUTES FOR A PERSON TO SPEND ON SOCIAL MEDIA IN A DAY ,\\
  IDEAL NUMBER OF TIMES FOR A PARENT TO PUNISH HIS/HER CHILD IN A MONTH ,\\
  IDEAL NUMBER OF MILES FOR A PERSON TO WALK IN A WEEK ,\\
  IDEAL PERCENTAGE OF PEOPLE IN ANY GIVEN CITY TO DRIVE DRUNK ,\\
  IDEAL NUMBER OF TIMES FOR A PERSON TO CHEAT ON A SIGNIFICANT OTHER IN A LIFETIME ,\\
  IDEAL NUMBER OF TIMES FOR A PERSON TO HIT SNOOZE ON AN ALARM CLOCK IN A DAY ,\\
  IDEAL NUMBER OF PARKING TICKETS FOR A PERSON TO RECEIVE IN AN YEAR ,\\
  IDEAL NUMBER OF TIMES FOR A PERSON TO GET HIS/HER CAR WASHED IN AN YEAR ,\\
  IDEAL NUMBER OF CUPS OF COFFEE FOR A PERSON TO DRINK IN A DAY ,\\
  IDEAL NUMBER OF DESSERTS FOR A PERSON TO CONSUME IN A WEEK ,\\
  IDEAL NUMBER OF LOADS OF LAUNDRY FOR A PERSON TO DO IN A WEEK ,\\
  IDEAL PERCENTAGE OF ADULTS IN ANY GIVEN CITY TO SMOKE ,\\
  IDEAL PERCENTAGE OF STUDENTS IN A HIGH SCHOOL TO DRINK UNDERAGE ,\\
  IDEAL PERCENTAGE OF PEOPLE TO LIE ON A DATING WEBSITE ,\\
  IDEAL NUMBER OF SERVINGS OF CARBOHYDRATES FOR A PERSON TO CONSUME IN A DAY ,\\
  IDEAL NUMBER OF TEXT MESSAGES FOR A PERSON TO SEND IN A DAY ,\\
  IDEAL NUMBER OF TIMES FOR A PERSON TO LOSE HIS/HER TEMPER IN A WEEK ,\\
  IDEAL NUMBER OF TIMES FOR A PERSON TO SWEAR IN A DAY \\

 \bottomrule
\end{tabular}
\caption{Experiment 2 ideal prompt}
\label{tab:exp1_app_prompts_idl}
 \end{table*}

\clearpage

\section{Case Study - Patient Recovery time}
\label{sec:exp1_case_study}

Results for the study shown from case study, showing negative aspects of a \textcolor{ideal}{prescriptive} norm when being misaligned with humans. The LLM is to predict recovery times for patients through its sample but instead of reporting its average recovery time, the sample returns one with a \textcolor{ideal}{prescriptive} component which is consistently lower than the average huring patient interests. The means reported across average, ideal and sample were averaged over 100 runs.
\mbox{}\\[0pt]
\begin{center}
\begin{minipage}{\dimexpr2\columnwidth+\columnsep\relax}
\centering
\small   
\begin{tabularx}{\textwidth}{|X|c|c|c|}
\hline
\textbf{Symptoms} & \textbf{Average} & \textbf{Ideal} & \textbf{Sample} \\
\hline
Increased thirst, Frequent urination, Fatigue, Blurred vision & 9.50 & 4.00 & 12.00 \\
\hline
Fever, Cough, Sore throat, Muscle aches & 2.50 & 2.30 & 2.50 \\
\hline
Wheezing, Shortness of breath, Chest tightness, Coughing, especially at night & 6.50 & 3.70 & 6.00 \\
\hline
Chronic cough, Mucus (sputum) production, Shortness of breath, Wheezing & 8.50 & 6.00 & 8.00 \\
\hline
Persistent cough, Weight loss, Night sweats, Fever & 10.50 & 10.00 & 10.00 \\
\hline
Chest pain (angina), Shortness of breath, Heart attack, Fatigue & 12.50 & 12.00 & 12.00 \\
\hline
Sudden numbness or weakness, Confusion or trouble speaking, Vision problems, Loss of balance or coordination & 12.50 & 12.00 & 12.00 \\
\hline
Tremors, Stiffness, Slowed movement, Balance problems & 12.50 & 12.00 & 12.10 \\
\hline
Joint pain, Swelling, Stiffness, Fatigue & 6.50 & 6.00 & 6.50 \\
\hline
Back pain, Loss of height over time, Stooped posture, Fractures & 12.40 & 12.00 & 12.00 \\
\hline
Fatigue, Weakness, Pale or yellowish skin, Shortness of breath & 5.30 & 4.60 & 6.50 \\
\hline
Diarrhea, Fatigue, Weight loss, Bloating and gas & 4.50 & 4.40 & 4.50 \\
\hline
Abdominal pain, Cramping, Bloating, Changes in bowel habits & 3.70 & 2.20 & 2.50 \\
\hline
Fever, Fatigue, Nausea and vomiting, Jaundice & 4.90 & 2.50 & 4.20 \\
\hline
Fever, Chills, Headache, Muscle pain & 2.50 & 2.00 & 2.40 \\
\hline
Fever, Rash, Joint pain, Red eyes & 2.50 & 2.10 & 2.10 \\
\hline
Skin sores, Numbness, Muscle weakness, Eye problems & 8.50 & 9.20 & 8.90 \\
\hline
Fever, Cough, Runny nose, Rash & 2.50 & 2.20 & 2.40 \\
\hline
Mild fever, Headache, Runny nose, Rash & 1.50 & 2.00 & 2.00 \\
\hline
Swollen, painful salivary glands, Fever, Headache, Muscle aches & 2.50 & 2.40 & 2.50 \\
\hline
Muscle stiffness, Muscle spasms, Difficulty swallowing, Fever & 6.50 & 4.30 & 5.30 \\
\hline
Fever, Headache, Excessive salivation, Muscle spasms & 4.50 & 3.10 & 3.70 \\
\hline
Severe cough, Whooping sound when inhaling, Vomiting, Exhaustion & 7.50 & 7.00 & 7.00 \\
\hline
Fever, Chills, Shortness of breath, Skin sores & 4.10 & 2.50 & 2.70 \\
\hline
Painless sores, Rash, Fever, Swollen lymph nodes & 3.90 & 4.00 & 4.00 \\
\hline
Painful urination, Abnormal discharge, Testicular pain, Pelvic pain & 4.50 & 2.50 & 2.50 \\
\hline
Painful urination, Abnormal discharge, Testicular pain, Pelvic pain & 4.50 & 2.50 & 2.50 \\
\hline
Genital warts, Itching, Discomfort, Bleeding with intercourse & 6.50 & 4.40 & 6.00 \\
\hline
Intense itching, Rash, Sores, Thick crusts on the skin & 2.50 & 2.80 & 3.40 \\
\hline
Red, itchy patches, Scaling, Blisters, Bald patches & 6.50 & 6.00 & 6.50 \\
\hline
Fatigue, Nausea, Jaundice, Dark urine & 6.50 & 6.00 & 6.10 \\
\hline
Stomach pain, Nausea, Vomiting, Bloating & 2.50 & 2.00 & 2.50 \\
\hline
Burning stomach pain, Bloating, Heartburn, Nausea & 3.30 & 2.00 & 3.60 \\
\hline
Sudden, intense pain in the abdomen, Nausea, Vomiting, Indigestion & 4.50 & 2.00 & 3.60 \\
\hline
\end{tabularx}
\captionof{table}{Experiment 2 Case Study - Patient Recovery time}
  \label{tab:exp1_patient_recovery}

\end{minipage}
\end{center}

\clearpage
\newpage
\section{Experiment 1 Glubbing experiment with other LLMs}
\label{sec:glubbing_other_LLM}
 We also check the presence of prescriptive norms replicating Experiment 1 in other LLMs. Results indicate that LLM sampling has a \textcolor{ideal}{prescriptive} and a \textcolor{average}{descriptive} component across a range of LLMs. The samples and the means reported were averaged over 100 runs.
\mbox{}\\[0pt]
\begin{center}
\begin{minipage}{\dimexpr2\columnwidth+\columnsep\relax} 
\tiny
\centering
\label{tab:glubbing_other_LLM}
\begin{tabularx}{\textwidth}{|X|X|X|X|}
\hline
\textbf{Model} & \textbf{Neg Ideal} & \textbf{Net Ideal} & \textbf{Pos Ideal} \\
\hline
\textbf{Llama-2-7b} & p-value: 0.000383 (Sig.) \newline $C_a$: 44.86, SD 1.65 \newline $C_s$: 36.80, SD 18.23 & p-value: 0.1159 (Not Sig.) \newline $C_a$: 45.15, SD 1.30 \newline $C_s$: 44.46, SD 18.38 & p-value: 0.6385 (Not Sig.) \newline $C_a$: 45.12, SD 1.67 \newline $C_s$: 46.13, SD 24.58 \\
\hline
\textbf{Llama-3-70b} & p-value: 0.0000875 (Sig.) \newline $C_a$: 44.96, SD 1.60 \newline $C_s$: 35.40, SD 17.21 & p-value: 0.560 (Not Sig.) \newline $C_a$: 45.10, SD 1.23 \newline $C_s$: 44.48, SD 16.33 & p-value: 0.000012 (Sig.) \newline $C_a$: 45.16, SD 1.47 \newline $C_s$: 46.68, SD 4.58 \\
\hline
\textbf{Mistral-7b} & p-value: 0.0543 (Not Sig.) \newline $C_a$: 45.23, SD 1.56 \newline $C_s$: 46.08, SD 5.39 & p-value: 0.7777 (Not Sig.) \newline $C_a$: 45.01, SD 1.43 \newline $C_s$: 44.24, SD 5.57 & p-value: 5.64e-17 (Sig.) \newline $C_a$: 44.96, SD 1.51 \newline $C_s$: 54.00, SD 4.83 \\
\hline
\textbf{Mixtral 8x7b} & p-value: 0.000708 ( Sig.) \newline $C_a$: 45.17, SD 1.86 \newline $C_s$: 46.86, SD 6.08 & p-value: 0.3094 (Not Sig.) \newline $C_a$: 45.14, SD 1.54 \newline $C_s$: 43.77, SD 8.08 & p-value: 1.80e-16 (Sig.) \newline $C_a$: 44.96, SD 1.49 \newline $C_s$: 54.17, SD 4.88 \\
\hline
\textbf{GPT-3.5} & p-value< 0.0001 ( Sig.) \newline $C_a$: 44.59, SD 1.84 \newline $C_s$: 37.31, SD 4.08 & p-value: 0.877 (Not Sig.) \newline $C_a$: 44.52, SD 1.52 \newline $C_s$: 44.92, SD 6.08 & p-value: 0.000021 (Sig.) \newline $C_a$: 44.84, SD 1.49 \newline $C_s$: 46.58, SD 4.68 \\
\hline
\textbf{GPT-4 (Temp 0)} & p-value< 0.0001 ( Sig.) \newline $C_a$: 44.80, SD 1.84 \newline $C_s$: 36.0, SD 2.02 & p-value: 0.913 (Not Sig.) \newline $C_a$: 44.73, SD 1.52 \newline $C_s$: 44.36, SD 2.03 & p-value< 0.0001 (Sig.) \newline $C_a$: 44.85, SD 1.48 \newline $C_s$: 46.58, SD 2.01 \\
\bottomrule
\end{tabularx}
\captionof{table}{Summary of Mann-Whitney U Test Results for Llama, Mistral, and Mixtral and GPT,showing significance in the majority of the cases}
\end{minipage}
\end{center}

\clearpage
\section{Experiment 3: List of prompts}
\label{sec:exp3_prompts}

\mbox{}\\[0pt]
\begin{center}
\begin{minipage}{\dimexpr2\columnwidth+\columnsep\relax}
    \centering
    \small
    \begin{tabular}{|p{.5cm}|p{.5cm}|p{13cm}|}
    \hline

        Cate-& Exem -& Passage \\
        gory & plar  &          \\ \hline
        1 & 1 & A 30-year-old woman who basically knows the material she is teaching, but is relatively uninspiring, boring to listen to, and not particularly fond of her job \\ 
        1 & 2 & A 25-year-old woman who captivates her students with exciting in-class demonstrations, grades assignments with remarkable speed, and inspires all of her students to succeed. Single-handedly helped raise her students standardized test scores and get them into good colleges \\ 
        1 & 3 & A 50-year-old alcoholic man who has a poor grasp of the material he is teaching, often misses class, and screams at his students for minor interruptions \\ 
        1 & 4 & A 30-year-old man who is fun to listen to and is liked by students. Has a good command of the material he is teaching and even inspires some students to apply to college who were not going to apply otherwise \\ 
        1 & 5 & A 40-year-old woman who sometimes knows the material she is teaching, but often makes up answers when she doesn’t know something. \\
        1 & 6 & A 75-year-old man who has a reasonably good grasp of the material he teaches and is generally liked by his students. Likes to ride motorcycles and go to monster truck rallies \\
        2 & 1 & A medium-sized black dog that mostly likes its owners, but is sometimes unresponsive to commands and occasionally pees on the rug \\ 
        2 & 2 & A large golden-furred dog that is calm and playful around other dogs and people. Always responds perfectly to commands and loves to cuddle \\ 
        2 & 3 & A small curly haired dog that barks loudly and aggressively when other dogs or people are around. Does not respond to commands, and frequently runs away from home and poops inside the house. Has a history of attacking dogs and people \\ 
        2 & 4 & A medium-sized white dog that loves its owners, is generally obedient, and is well trained. Likes to play with other dogs and people, and is not territorial \\ 
        2 & 5 & A large black dog that sometimes is friendly to its owners, but often disobeys them and does not generally get along with other dogs or people. Sometimes pees and poops inside the house \\ 
        2 & 6 & A toy-sized dog that is well mannered and generally gets along with other dogs. Its fur is purple, and it has gigantic ears. Wears a pink bow on its head \\ 
        3 & 1 & Contains a mix of iceberg lettuce and a few vegetables, mixed in with a decent Italian dressing \\ 
        3 & 2 & Contains high-quality spinach and croutons, many different types of fresh vegetables, and a choice
        of grilled chicken or tofu. Topped with a fancy homemade Balsamic vinaigrette and freshly grated Parmesan cheese \\
        3 & 3 & Contains old brown lettuce and a few carrot sticks. Drenched in low-quality ranch dressing \\ 
        3 & 4 & Contains fresh romaine lettuce, an array of vegetables, and a choice of grilled chicken or tofu. Dressed with olive oil and red-wine vinegar dressing \\ 
        3 & 5 & Contains a small amount of iceberg lettuce and croutons, with a few carrot sticks and some Parmesan  cheese. Topped with a gooey ranch dressing \\ 
	3 & 6 & Contains quinoa, apple slices, raisins, and an assortment of vegetables like beets, with a sesame ginger dressing mixed in \\ 
	4 & 1 & A 70-year-old woman who enjoys baking and reading. Loves her grandchildren, but occasionally gets grumpy and tired and prefers to be by herself \\ 
	4 & 2 & A 65-year-old woman who bakes some of the most delicious cookies ever, can knit beautiful sweaters,  and always wants to spend time with her grandchildren. Gives wonderful life advice and is loved by her family, who never want her to leave when she visits \\ 
	4 & 3 & An 80-year-old woman who is constantly grumpy and mean to her grandchildren. Detests spending time with other people, but always demands that her children do favors for her. Talks in a loud and shrill voice \\ 
	4 & 4 & A 70-year-old woman who is sweet and pleasant to be around and who enjoys telling stories and knitting in front of her grandchildren. Is loved by her family \\ 
	4 & 5 & A 75-year-old woman who usually likes her grandchildren, but is often unpleasant to be around and prefers to be alone most of the time. Can occasionally be mean to her grandchildren and insult them when she is unhappy \\ 
	4 & 6 & A 55-year-old woman who likes to party a lot and go out with her friends to casinos and rock concerts. Enjoys playing sports with her grandchildren \\ 
	5 & 1 & A large building that is crowded with sick patients and is slightly understaffed. The nurses keep accurate records and are generally in control of things, but wait times, especially in the emergency room, tend to be long \\

 \bottomrule

    \end{tabular}
    \captionof{table}{List of passages used in Experiment 3, each row consists of a concept and an exemplar of that concept along with the passage. These passages are rated along three dimensions of: average, ideal and protypicality}
    \label{tab:exp3_app_passages}

\end{minipage}
\end{center}

\clearpage
\newpage

\mbox{}\\[0pt]
\begin{center}
\begin{minipage}{\dimexpr2\columnwidth+\columnsep\relax}
    \centering
    \small
    \begin{tabular}{|p{.5cm}|p{.5cm}|p{13cm}|}
    \hline

        Cate-& Exem -& Passage \\
        gory & plar  &          \\ \hline
    5 & 2 & A pristine building in a quiet, beautiful area overlooking the mountains. Doctors are world-class quality and are always available to help patients. Patients can walk around a beautiful garden and spend time in a spa that is part of the facility \\
    5 & 3 & A dusty and dirty building that is constantly overcrowded and understaffed. Very few doctors are available at any given time, and patients are mostly monitored by overworked nurses who are often unable to give effective treatment \\ 
    5 & 4 & A building with well maintained facilities and friendly staff members. Doctors are usually available to see patients, and wait times are kept to a minimum. Patients report receiving good treatment \\ 
    5 & 5 & An ugly building with old facilities. Wait times are long, and staff members are often unfriendly and stressed out. Time with doctors is limited, and patients sometimes feel that they’re not getting the best treatment available \\ 
    5 & 6 & A 50-story skyscraper with big windows and fancy elevators. Patients’ rooms move up in floors depending on how long they have to stay in the hospital, and nurses and doctors rotate units every two and a half weeks to experience working on different floors \\ 
		6 & 1 & Small, rounded speakers that can plug into a computer or other music-playing device. Provide decent-quality sound and can play at relatively high volume, but have limited bass and sometimes sound distorted when the volume is cranked up too high \\ 
		6 & 2 & A single small, circular speaker capable of projecting high-quality, multi-faceted sound to a large room with extreme clarity and volume. Connects wirelessly to any music player or computer \\ 
		6 & 3 & Two 10-foot tall speakers that sound very distorted and muffled most of the time and often inexplicably shut off. Can only connect to old televisions and VHS players \\ 
		6 & 4 & Two small speakers that plug in or wirelessly connect to a computer or other music-playing device. Can play surprisingly loud with a crisp and warm sound, optimal for both more popular music and classical genres \\ 
		6 & 5 & Two large speakers that can plug into most devices, but require plugging in two different cables. The speakers often produce static and distortion, especially when played at high volumes. Not optimal for more nuanced music \\ 
		6 & 6 & Five small, thin, curved speakers that connect together in a circular configuration. Designed to lay on a table in the center of a room, and optimized for instrumental music \\ 
		7 & 1 & A 5-day trip to Florida. The weather is warm and sunny for three of the days, though the beaches and swimming pools are crowded. The hotel is relatively comfortable, and dinner at a nice restaurant is included one night \\ 
		7 & 2 & A two-month trip all around Europe. Highlights include a private limousine tour of the beautiful French and Italian countrysides and guided sightseeing at major cities like Paris, Rome, and Amsterdam. Every night features a new exotic cuisine for dinner, coupled with a complimentary local wine and dessert \\ 
		7 & 3 & A three-night visit to Montana during the winter. The weather is very cold, and the motel room is musty and cramped. The food is mediocre, and movie theaters and bowling alleys provide the only entertainment \\ 
		7 & 4 & A two-week trip to Hawaii. Includes tours of the volcanoes and vacationing on the beach. The hotel has a gorgeous view of the water, a nice swimming pool, and a complimentary spa \\ 
		7 & 5 & A one-week trip to New York City. The weather is mostly cold and rainy, and the hotel is old and smelly. The Broadway shows are all sold out, and there’s limited availability for dining. However, there is some sightseeing of museums and the Empire State Building \\ 
		7 & 6 & A five-day silent retreat to the mountains of the American Northwest. Most of the days are spent hiking and meditating. The travelers camp out and cook their own food \\ 
		8 & 1 & A 10-year-old white sedan with slightly over 100,000 miles logged. Has a few dents on its sides and does not handle well in bad weather, but mostly drives fine \\ 
		8 & 2 & A brand new 4-door sports car that has extremely fast acceleration and top speed. Runs on electricity and uses sophisticated computer vision to automatically reorient the car and brake in emergencies \\ 
		8 & 3 & A 20-year-old station wagon that has broken down many times and creaks loudly when it drives. Sometimes the ignition doesn’t work, and the car doesn’t start. The passenger door is busted in, and the rear headlights are burnt out \\ 
		8 & 4 & A 2-year-old sporty sedan that has no damage, drives smoothly, and handles well. Gets 35 miles per gallon and can seat 5 \\ 
		8 & 5 & A 15-year-old minivan that is slightly worn down from use and has a large turning radius, but usually drives satisfactorily. Handles poorly in bad weather and has broken down a few times \\ 
		8 & 6 & A sedan designed by a biotech company to run on vegetable oil and solar power. The car recycles its own energy to provide heat and air conditioning \\ \hline
    \end{tabular}
    \captionof{table}{List of passages used in Experiment 3, each row consists of a concept and an exemplar of that concept along with the passage. These passages are rated along three dimensions of: average, ideal and protypicality}
\end{minipage}
\end{center}

\clearpage
\section{Experiment 3 complete results}
\label{sec:exp3_results}
\mbox{}\\[0pt]
\begin{center}
\begin{minipage}{\dimexpr2\columnwidth+\columnsep\relax}
  \centering
  \small
  \resizebox{\textwidth}{!}{%
  \begin{tabular}{l l l l l l l l}
    \toprule
    concept Code & Exemplar Code & Average & Ideal & Good Example & Paradigm Example & Proto. Example & Composite \\
    \hline
    1.00 & 1.00 & 4.50 & 2.00 & 2.50 & 4.50 & 4.50 & 3.83 \\
1.00 & 2.00 & 1.00 & 7.00 & 7.00 & 6.50 & 6.50 & 6.67 \\
1.00 & 3.00 & 0.50 & 0.00 & 0.00 & 0.50 & 0.50 & 0.33 \\
1.00 & 4.00 & 4.50 & 7.00 & 7.00 & 6.50 & 6.50 & 6.67 \\
1.00 & 5.00 & 3.50 & 0.50 & 1.50 & 1.50 & 1.50 & 1.50 \\
1.00 & 6.00 & 2.50 & 5.50 & 5.50 & 4.50 & 2.50 & 4.17 \\
2.00 & 1.00 & 5.50 & 3.50 & 5.50 & 4.50 & 4.50 & 4.83 \\
2.00 & 2.00 & 4.50 & 7.00 & 7.00 & 6.50 & 6.50 & 6.67 \\
2.00 & 3.00 & 0.50 & 0.00 & 1.50 & 1.50 & 1.00 & 1.33 \\
2.00 & 4.00 & 5.50 & 6.50 & 6.50 & 6.50 & 6.50 & 6.50 \\
2.00 & 5.00 & 2.50 & 1.50 & 2.50 & 2.50 & 2.50 & 2.50 \\
2.00 & 6.00 & 0.00 & 4.50 & 1.50 & 1.50 & 1.00 & 1.33 \\
3.00 & 1.00 & 6.50 & 4.50 & 5.50 & 6.50 & 6.50 & 6.17 \\
3.00 & 2.00 & 4.50 & 6.50 & 6.50 & 6.50 & 6.50 & 6.50 \\
3.00 & 3.00 & 2.50 & 0.50 & 1.50 & 2.50 & 2.50 & 2.17 \\
3.00 & 4.00 & 5.50 & 5.50 & 6.50 & 6.50 & 6.50 & 6.50 \\
3.00 & 5.00 & 5.50 & 4.50 & 5.50 & 5.50 & 5.50 & 5.50 \\
3.00 & 6.00 & 2.50 & 5.50 & 6.50 & 5.50 & 5.50 & 5.83 \\
4.00 & 1.00 & 6.50 & 5.50 & 6.50 & 6.50 & 6.50 & 6.50 \\
4.00 & 2.00 & 5.50 & 7.00 & 7.00 & 7.00 & 7.00 & 7.00 \\
4.00 & 3.00 & 1.50 & 0.50 & 0.50 & 1.50 & 1.50 & 1.17 \\
4.00 & 4.00 & 5.50 & 7.00 & 7.00 & 7.00 & 6.50 & 6.83 \\
4.00 & 5.00 & 3.50 & 2.50 & 2.50 & 2.50 & 2.50 & 2.50 \\
4.00 & 6.00 & 2.50 & 5.50 & 5.50 & 4.50 & 3.50 & 4.50 \\
5.00 & 1.00 & 5.50 & 2.50 & 5.50 & 5.50 & 5.50 & 5.50 \\
5.00 & 2.00 & 0.50 & 7.00 & 5.50 & 2.50 & 2.50 & 3.50 \\
5.00 & 3.00 & 1.50 & 0.00 & 0.50 & 1.50 & 1.50 & 1.17 \\
5.00 & 4.00 & 5.50 & 7.00 & 6.50 & 6.50 & 6.50 & 6.50 \\
5.00 & 5.00 & 4.50 & 0.00 & 1.50 & 4.50 & 2.50 & 2.83 \\
5.00 & 6.00 & 0.00 & 4.50 & 2.50 & 1.50 & 1.50 & 1.83 \\
6.00 & 1.00 & 5.50 & 4.50 & 4.50 & 4.50 & 4.50 & 4.50 \\
6.00 & 2.00 & 1.50 & 6.50 & 2.50 & 4.50 & 4.50 & 3.83 \\
6.00 & 3.00 & 0.00 & 0.50 & 0.50 & 0.50 & 0.50 & 0.50 \\
6.00 & 4.00 & 5.50 & 6.50 & 6.50 & 6.50 & 6.50 & 6.50 \\
6.00 & 5.00 & 4.50 & 1.50 & 3.50 & 4.50 & 4.50 & 4.17 \\
6.00 & 6.00 & 0.50 & 5.50 & 2.50 & 2.50 & 1.50 & 2.17 \\
7.00 & 1.00 & 5.50 & 5.50 & 5.50 & 6.50 & 6.50 & 6.17 \\
7.00 & 2.00 & 0.00 & 7.00 & 7.00 & 6.50 & 5.50 & 6.33 \\
7.00 & 3.00 & 4.50 & 1.50 & 1.50 & 1.50 & 1.50 & 1.50 \\
7.00 & 4.00 & 2.50 & 6.50 & 6.50 & 6.50 & 6.50 & 6.50 \\
7.00 & 5.00 & 4.50 & 2.50 & 2.50 & 3.50 & 3.50 & 3.17 \\
7.00 & 6.00 & 1.50 & 5.50 & 5.50 & 4.50 & 2.50 & 4.17 \\
8.00 & 1.00 & 5.50 & 2.50 & 4.50 & 4.50 & 4.50 & 4.50 \\
8.00 & 2.00 & 0.50 & 6.50 & 6.50 & 6.50 & 4.50 & 5.83 \\
8.00 & 3.00 & 0.50 & 0.00 & 0.50 & 1.50 & 1.50 & 1.17 \\
8.00 & 4.00 & 5.50 & 6.50 & 6.50 & 6.50 & 6.50 & 6.50 \\
8.00 & 5.00 & 3.50 & 2.50 & 3.50 & 3.50 & 3.50 & 3.50 \\
8.00 & 6.00 & 0.00 & 6.50 & 6.50 & 1.50 & 1.50 & 3.17 \\

     \bottomrule
  \end{tabular}}
  \captionof{table}{Experiment 3 results based on how the LLM rates prototypes on three dimensions namely, average, ideal and protypicality. Prototypicality is further subdivided into 3 types, of being a good example, a paradigm example and a prototypical example, composite score is the average across the three prototypicality scores }
  \label{tab:exp3_app_values}

\end{minipage}
\end{center}

\onecolumn

\section{Full List of concepts}

\label{full_cat}
\begin{longtable}{|p{5cm}|p{10cm}|}

\hline
\textbf{Category} & \textbf{Concepts} \\ \hline
\endfirsthead

\hline
\textbf{Category} & \textbf{Concepts} \\ \hline
\endhead

\hline
\endfoot

\textbf{Education, childcare and school} & 
Percentage of students in a middle school to be bullied \newline
Percentage of students in a high school to dropout \newline
Percentage of students in a high school to cheat on an exam \newline
Number of times for a parent to punish child in a month \newline
Percentage of students in a high school to drink underage \newline
Number of extracurricular activities a student participates in a school year \newline
Number of complaints received about school bus behavior in a year \newline
Percentage of students failing a subject in a school year \newline
Percentage of high school students participating in sports \newline
Number of hours students spend on homework in middle school \newline
Number of parent-teacher meetings a parent attends in a school year \newline
Number of conflicts between parents and school staff in a year \newline
Number of field trips students attend per school year \newline
Number of fire or safety incidents reported at school in a year \newline
Number of hours a child uses digital devices for learning purposes in a day \newline
Percentage of students in a middle school using a school library daily \newline
Number of science fair projects a student completes in a school year \newline
Percentage of high school students involved in a student government \newline
Number of times a student is late to school in a month \newline
Percentage of students completing advanced placement courses in high school \newline
Number of school assemblies a student attends in a year \newline
Percentage of students volunteering for community service through school programs \newline
Percentage of students in elementary school walking to school \newline
Percentage of students with perfect attendance records in a school year \newline
Number of art projects completed by a student in a school year. \\ \hline

\textbf{Urban social statistics} & 
Number of graffiti incidents reported in a city in a month \newline
Percentage of people in a city who jaywalk in a week \newline
Number of noise complaints filed in a neighborhood in a month \newline
Percentage of city residents who use public transportation daily \newline
Number of times residents participate in community clean-up events in a year \newline
Percentage of people in a city who participate in local elections \newline
Number of public disturbances reported in a city in a month \newline
Percentage of residents involved in neighborhood disputes in a year \newline
Number of times a person uses a car-sharing service in a month \newline
Percentage of residents who recycle regularly in a city \newline
Number of stray animals reported in urban areas in a month \newline
Percentage of city residents who volunteer for social services in a year \newline
Number of times to litter in public spaces in a month \newline
Percentage of residents living below the poverty line in a city \newline
Number of public intoxication arrests in a city in a year \newline
Number of parking tickets to receive in a year \newline
Number of times to swear in a day \newline
Number of times to honk at other drivers in a week \newline
Percentage of people in any city to drive drunk \newline
Percentage of adults in any city to smoke \newline
Number of times to report a lost or found item in a city in a year \newline
Percentage of residents who use bikes as their primary mode of transportation in a city \newline
Number of illegal parking incidents reported in a city in a month \newline
Percentage of people using ride-sharing apps in urban areas on a daily basis \newline
Number of times residents complain about public transport delays in a month \newline
Percentage of urban residents owning pets. \\ \hline

\textbf{Health and fitness} & 
Number of sugary drinks to consume in a week \newline
Number of hours to spend exercising in a week \newline
Number of calories to consume in a day \newline
Number of miles to walk in a week \newline
Number of servings of carbohydrates to consume in a day \newline
Number of hours to sleep in a night \newline
Number of desserts to consume in a week \newline
Number of cups of coffee to drink in a day \newline
Number of times to visit a doctor for routine check-ups in a year \newline
Number of minutes to spend meditating in a day \newline
Number of days per week to engage in strength training exercises \newline
Number of servings of protein to consume in a day \newline
Number of glasses of water to drink in a day \newline
Number of fast food meals to consume in a week \newline
Number of times to use a standing desk instead of sitting in a week \newline
Number of hours of screen time in a day \newline
Number of steps to take in a day \newline
Number of alcoholic beverages to consume in a week \newline
Number of times to apply sunscreen before going outdoors in a week \newline
Number of minutes to spend stretching in a day \newline
Number of servings of leafy greens to consume in a day \newline
Number of minutes to spend in direct sunlight in a day \newline
Number of health apps to used for tracking fitness or diet \newline
Number of weight measurements to take in a month \newline
Number of times to consult a nutritionist or dietitian in a year \newline
Number of dental check-ups to schedule in a year. \\ \hline

\textbf{Social media and internet usage} & 
Number of times to call parents in a month \newline
Number of minutes to spend on social media in a day \newline
Number of text messages to send in a day \newline
Number of times to check emails in a day \newline
Number of times to post on social media platforms in a week \newline
Number of hours to spend watching streaming services in a day \newline
Number of online shopping sessions in a month \newline
Number of online courses to enroll in per year \newline
Number of online games to play in a week \newline
Number of times to back up digital data in a month \newline
Number of times to clear browsing history and cookies in a month \newline
Number of podcasts to listen to in a week \newline
Number of new online friends or contacts added in a month \newline
Number of apps downloaded in a month \newline
Number of times to participate in virtual meetings in a week \newline
Number of online petitions signed in a year \newline
Number of times to change main online passwords in a year \newline
Percentage of daily internet use for educational purposes \newline
Times a user changes their main profile photo on social media in a year \newline
Number of unique social media platforms visited in a week \newline
Number of online accounts deactivated or closed each year \newline
Frequency of using private or incognito browsing modes each week \newline
Frequency of checking news websites daily \newline
Monthly instances of donating to online fundraisers or charity drives \newline
Number of ad blockers installed or active on devices each month \newline
Frequency of commenting on blogs or online articles each week. \\ \hline

\textbf{Habits, behavior and lifestyle} & 
Number of hours of TV to watch in a day \newline
Number of servings of fruits and vegetables to consume in a month \newline
Number of lies to tell in a week \newline
Number of times to check phone in a day \newline
Number of romantic partners to have in a lifetime \newline
Number of books to read in a year \newline
Percentage of people to lie on a dating website \newline
Number of times to lose temper in a week \newline
Number of times to clean home in a month \newline
Number of times to hit snooze on an alarm clock in a day \newline
Number of times to get car washed in a year \newline
Number of loads of laundry to do in a week \newline
Number of times to visit a museum or cultural event in a year \newline
Number of family meals to have per week \newline
Number of plants to care for in the home \newline
Number of new skills or hobbies to start learning each year \newline
Number of social events attended each month \newline
Number of health check-ups scheduled annually \newline
Number of meals cooked at home each week \newline
Number of times to change bed linens in a month \newline
Number of days per week dedicated to device-free time \newline
Percentage of clothing purchases that are from sustainable brands each year \newline
Number of cups of water to drink in a day \newline
Number of personal emails to send in a week \newline
Number of hours to listen to music in a day \newline
Number of journal entries to write in a month. \\ \hline

\textbf{Wealth and Economic habits} & 
Dollars of tax evaded by a person in a year \newline
Number of credit cards owned by a person \newline
Percentage of income saved annually \newline
Number of times a person shops online in a month \newline
Amount of money spent on dining out in a month \newline
Number of times a person checks their bank account balance in a week \newline
Number of loans taken out in a lifetime \newline
Dollars spent on impulse purchases in a month \newline
Dollars spent for buying electronics in an year \newline
Percentage of salary spent on housing \newline
Dollars of total saving in a year \newline
Number of luxury items purchased in a year \newline
Amount of money donated to charity annually \newline
Number of times a person reviews their budget in a month \newline
Percentage of income spent on entertainment \newline
Number of times a person consults a financial advisor in a year \newline
Amount of debt carried by a person on average \newline
Number of times a person uses a coupon in a month \newline
Amount of emergency savings recommended for a person \newline
Number of investment accounts owned \newline
Percentage of income spent on travel annually \newline
Number of times a person revises their will in a lifetime \newline
Number of financial seminars or workshops attended in a year \newline
Amount of money spent on subscriptions in a month \newline
Number of times a person renegotiates their salary in a career \newline
Number of times a person invests in stocks in a month. \\ \hline

\textbf{Environmental Sustainability} & 
Number of trees planted by a person in a year \newline
Number of times a person uses a reusable shopping bag in a month \newline
Amount of water saved by using water-efficient fixtures in a year \newline
Number of days a person participates in carpooling in a month \newline
Amount of energy saved by using energy-efficient appliances in a year \newline
Number of plastic bottles recycled by a person in a month \newline
Percentage of household waste composted \newline
Number of times a person rides a bicycle instead of driving in a week \newline
Amount of food waste reduced by a person in a month \newline
Number of times a person participates in community clean-up events in a year \newline
Percentage of products purchased that are made from recycled materials \newline
Number of times a person uses public transportation in a week \newline
Amount of greenhouse gas emissions reduced by using renewable energy sources in a year \newline
Percentage of clothing purchased that is second-hand or sustainably made \newline
Number of times a person participates in environmental advocacy or activism in a year \newline
Number of times a person chooses eco-friendly packaging options in a month \newline
Percentage of cleaning products used that are eco-friendly \newline
Number of times a person opts for plant-based meals in a week \newline
Amount of money spent on supporting environmental causes in a year \newline
Number of times a person uses single-use plastic in a week \newline
Amount of food waste thrown away in a month \newline
Number of times a person leaves lights on in empty rooms in a day \newline
Number of disposable coffee cups used in a month \newline
Amount of water wasted by leaving taps running in a month \newline
Amount of fuel wasted by idling a car in a week \newline
Number of times a person fails to separate recyclables from regular trash in a month. \\ \hline

\textbf{Politics and international relationships} & 
Number of international conflicts in a year \newline
Number of treaties or agreements signed by a country in a year \newline
Number of times a person votes in national elections in a lifetime \newline
Number of diplomatic visits made by a country's leaders in a year \newline
Percentage of a country's budget allocated to defense spending \newline
Number of international organizations a country is a member of \newline
Number of international trade agreements signed in a year \newline
Percentage of foreign aid given by a country as a portion of GDP \newline
Number of times a person participates in political protests in a year \newline
Number of bilateral meetings held between countries in a year \newline
Number of sanctions imposed by a country in a year \newline
Percentage of citizens who support international cooperation \newline
Number of diplomatic embassies a country maintains worldwide \newline
Number of refugees accepted by a country in a year \newline
Number of international espionage incidents reported in a year \newline
Number of military bases a country has abroad \newline
Percentage of international agreements ratified by a country's parliament \newline
Number of international cultural exchange programs sponsored in a year \newline
Number of cyberattacks attributed to foreign governments in a year \newline
Number of international humanitarian missions a country participates in a year \newline
Number of trade disputes resolved through international arbitration in a year \newline
Number of international human rights organizations criticizing a country's policies in a year \newline
Number of times a country is accused of violating international law in a year \newline
Number of military conflicts a country initiates in a year \newline
Number of times a country faces international boycotts due to its policies in a year \newline
Percentage of the population living under undemocratic regimes. \\ \hline

\textbf{Technology and Innovation} & 
Number of smartphone models that sold more than 10,000 pieces in a year \newline
Average number of hours people spend on social media per day \newline
Number of new technology products introduced to the market in a year \newline
Average age at which people purchase their first smartphone \newline
Percentage of households with smart home devices \newline
Average number of apps installed on a smartphone \newline
Number of electric vehicles sold in a country in a year \newline
Average number of hours people spend on online gaming per week \newline
Percentage of households with high-speed internet access \newline
Number of people using wearable fitness trackers in a country \newline
Average lifespan of a smartphone before being replaced \newline
Percentage of people using online banking services \newline
Number of streaming service subscriptions per household \newline
Average number of data breaches affecting consumers per year \newline
Percentage of consumers using mobile payment systems \newline
Average number of times people upgrade their tech devices in a year \newline
Number of people using telemedicine services in a country per year \newline
Percentage of market share held by electric vehicles \newline
Average amount of money spent by consumers on new technology annually \newline
Number of electric vehicle charging stations installed in a country per year \newline
Average number of hours people spend on virtual reality per week \newline
Percentage of consumers purchasing technology products online \newline
Number of broadband internet subscribers in a country \newline
Average number of new apps downloaded per person per year \newline
Number of households using renewable energy technology. \\ \hline

\textbf{Pet Care and Ownership} & 
Number of animals rescued and adopted in a year \newline
Average number of pets owned per household \newline
Amount of money spent on pet food annually \newline
Number of veterinary visits per pet per year \newline
Percentage of households with at least one pet \newline
Number of pet grooming sessions per year \newline
Amount of money spent on pet healthcare annually \newline
Number of pet-related products purchased per month \newline
Percentage of pets that are spayed or neutered \newline
Average lifespan of different pet species \newline
Number of times a pet is walked per day \newline
Amount of money spent on pet toys annually \newline
Number of pet-friendly parks or areas in a city \newline
Percentage of pets with microchips \newline
Number of pet training sessions attended per year \newline
Amount of money spent on pet insurance annually \newline
Number of pets abandoned or surrendered per year \newline
Percentage of pet owners who travel with their pets \newline
Number of pet-related accidents or injuries per year \newline
Average cost of pet adoption fees \newline
Percentage of households with multiple pets \newline
Number of pet-related events or expos attended per year \newline
Amount of money spent on pet boarding or daycare annually \newline
Number of pet adoptions from shelters versus breeders \newline
Percentage of pet owners who feed their pets homemade food. \\ \hline

\textbf{Travel, Tourism and Hospitality} & 
Number of countries visited by a person in their lifetime \newline
Average number of vacations taken per year \newline
Percentage of vacations that are international trips \newline
Number of cultural or heritage sites visited per year \newline
Average amount of money spent on travel annually in dollars \newline
Number of luxury cruises taken in a lifetime \newline
Percentage of travel done for leisure versus business \newline
Number of times a person stays at eco-friendly accommodations per year \newline
Average duration of an international trip in days \newline
Number of languages a person learns basic phrases of for travel \newline
Number of travel blogs or reviews written by a person in a lifetime \newline
Number of adventure or extreme sports tried while traveling \newline
Average number of travel souvenirs collected per trip \newline
Percentage of travel plans made spontaneously versus planned in advance \newline
Number of times a person travels with family per year \newline
Number of times a person visits the same destination multiple times \newline
Number of travel cancellations or delays experienced in a year \newline
Amount of money lost due to travel scams or fraud in a lifetime \newline
Number of times a person experiences food poisoning while traveling \newline
Number of travel insurance claims filed in a year \newline
Percentage of vacations that end with dissatisfaction or complaints \newline
Number of countries visited where a person experiences significant cultural differences \newline
Number of travel destinations visited due to trending social media recommendations \newline
Number of times a person misses a flight or train in a lifetime \newline
Amount of money spent on unexpected travel expenses annually \newline
Number of positive travel reviews written in a year. \\ \hline

\end{longtable}

\end{document}